\documentclass[lettersize,journal]{IEEEtran}
\usepackage{amsmath,amsfonts}
\usepackage{algorithmic}
\usepackage{algorithm}
\usepackage{array}
\usepackage[caption=false,font=normalsize,labelfont=sf,textfont=sf]{subfig}
\usepackage{textcomp}
\usepackage{stfloats}
\usepackage{url}
\usepackage{verbatim}
\usepackage{graphicx}
\usepackage{cite}
\usepackage{multirow}
\usepackage{graphicx}
\begin{document}

\title{Human Attention-Guided Explainable Artificial Intelligence for Computer Vision Models}

\author{Guoyang Liu, Jindi Zhang, Antoni B. Chan, Janet H. Hsiao
        % <-this % stops a space
\thanks{We are grateful to Huawei and Research Grants Council of Hong Kong (Collaborative Research Fund No. C7129-20G to Dr. J. Hsiao). We also thank Yumeng Yang and Yueyuan Zheng for their help in data collection. (Corresponding author: Janet H. Hsiao).}% <-this % stops a space        
\thanks{Guoyang Liu is with the Department of Psychology, University of Hong Kong, Pokfulam Road, Hong Kong (e-mail: gyangliu@hku.hk).}% <-this % stops a space
\thanks{Jindi Zhang is with Huawei Research Hong Kong (e-mail: zhangjindi2@huawei.com).}% <-this % stops a space
\thanks{Antoni B. Chan is with Department of Computer Science, City University of Hong Kong, Kowloon Tong, Hong Kong (email: abchan@cityu.edu.hk).}% <-this % stops a space
\thanks{Janet H. Hsiao is with the Department of Psychology, the State Key Laboratory of Brain and Cognitive Sciences, and the Institute of Data Science, University of Hong Kong, Pokfulam Road, Hong Kong (email: jhsiao@hku.hk).}% <-this % stops a space

\thanks{Manuscript received xxx xx, 20xx; revised xxx xx, 20xx.}}

\markboth{Journal of \LaTeX\ Class Files,~Vol.~14, No.~8, August~2021}%
{Shell \MakeLowercase{\textit{et al.}}: A Sample Article Using IEEEtran.cls for IEEE Journals}

% % column for its text to clear the IEEEpubid mark.

\maketitle

\begin{abstract}
We examined whether embedding human attention knowledge into saliency-based explainable AI (XAI) methods for computer vision models could enhance their plausibility and faithfulness.  We first developed new gradient-based XAI methods for object detection models to generate object-specific explanations by extending the current methods for image classification models. Interestingly, while these gradient-based methods worked well for explaining image classification models, when being used for explaining object detection models, the resulting saliency maps generally had lower faithfulness than human attention maps when performing the same task. We then developed Human Attention-Guided XAI (HAG-XAI) to learn from human attention how to best combine explanatory information from the models to enhance explanation plausibility by using trainable activation functions and smoothing kernels to maximize XAI saliency map’s similarity to human attention maps. While for image classification models, HAG-XAI enhanced explanation plausibility at the expense of faithfulness, for object detection models it enhanced plausibility and faithfulness simultaneously and outperformed existing methods. The learned functions were model-specific, well generalizable to other databases.  
\end{abstract}

\begin{IEEEkeywords}
Object detection, XAI, human attention, deep learning, saliency map.
\end{IEEEkeywords}

\section{Introduction}
\IEEEPARstart{I}{n} the last decades, deep learning technology has tremendously developed and revolutionized the field of artificial intelligence (AI) \cite{RN18}. Nowadays, deep learning has been widely applied in image classification, object detection, and natural language processing applications \cite{RN4}. However, the black-box nature and high computational complexity of deep learning models have made their decision-making process opaque to users, significantly affecting user trust and their usefulness \cite{RN31}. Many explainable AI (XAI) methods thus have been proposed to help human users comprehend AI decision making processes and enhance their trust of the output produced by AI. A common XAI method for computer vision models has been to use a saliency map to highlight features contributing to AI systems’ decisions. Current saliency-based XAI methods for image classification can be roughly classified into two categories: gradient-based and perturbation-based \cite{RN2}. Grad-CAM \cite{RN33} and Grad-CAM++ \cite{RN7} are two representative gradient-based XAI methods for image classification models. While they work well for image classification models, they are not suitable for object detection models because they can only generate class-specific rather than object-specific saliency maps. RISE \cite{RN25} and DRISE \cite{RN26} are two widely used perturbation-based XAI methods, which infer input features contributing to model decisions by perturbing the model input and observing the corresponding output changes. Similarly, although this method is shown to work well for image classification models, it  does not work well for object detection models because a large number of masks of different sizes are required for detecting multiple target instances of different sizes. Thus, this method is computationally expensive and often results in noisy backgrounds in the saliency map that are not interpretable, significantly impacting its explanation performance \cite{RN38, RN19}. 

Saliency-based XAI methods are typically evaluated on two main aspects, faithfulness and plausibility. Faithfulness measures how well the highlighted regions of a saliency map reflect features diagnostic to AI’s decisions \cite{RN7, RN32}. It is typically assessed by examining the amount of AI model’s performance change resulting from deletion or insertion of highlighted features. Plausibility measures whether the interpretations of AI’s operations conform to human cognition \cite{RN36}. It was typically measured by subjective human judgments \cite{RN7}.  Current saliency-based XAI methods for image classification, including Grad-CAM and Grad-CAM++, have been reported to achieve good faithfulness and plausibility \cite{RN3, RN7}. In contrast, in object detection, although existing XAI methods can roughly locate important regions for the decision of AI, the low spatial resolution of gradient-based methods and the dispersion problem intrinsic to perturbation-based methods seriously affect their faithfulness and plausibility \cite{RN19}. Object detection plays an important role in many critical AI systems, such as autonomous driving \cite{RN1, RN5, RN8} and medical diagnosis \cite{RN3, RN22}. These limitations make current XAI methods hard to be applied to these high safety-demanded control systems \cite{RN24}. Thus, it is essential to develop effective XAI methods for object detection models with better faithfulness and plausibility, aiming to address the above limitations, to make them more useful and accessible to users.

In recent years, human-in-the-loop approaches have been proposed to enhance the interpretability of machine learning models \cite{RN44}. More specifically, human attention information has been integrated into the development of AI models to evaluate or enhance AI’s performance and interpretability, such as to guide information search in some visual question answering (VQA) applications \cite{RN45, RN46}, or to enhance predictions from deep attention neural networks \cite{RN47, RN49, RN48}. Recent studies on XAI for image classification \cite{RN43} and VQA \cite{RN46} also indicated that human-like explanations could make users trust in the system. Nevertheless, XAI methods for computer vision models, especially for object detection models, which leverage human attention information to guide the generation of explanations, remains very limited. Human attention reflects information processing mechanisms underlying human cognition \cite{RN14}. In tasks that are commonly performed by computer vision models such as image classification and object detection, human attention when performing the same tasks, or when providing explanations for these tasks, highlights features essential for human information processing in visual cognition or explanation \cite{RN28, RN29, RN34, RN12}. Thus, it can potentially help guide XAI methods to provide explanations that are more accessible to humans.

Most recently, human attention has also been used to provide an objective measure of plausibility for saliency-based XAI methods by calculating their similarity to human attention maps \cite{RN23, RN35}, with the assumption that an explanation is good if it correlates with human attention deployed in visual tasks \cite{RN50}. Since collecting human data for generating human attention maps is often time-consuming and labor-intensive, Yang et al. \cite{RN35} proposed a human saliency imitator (HSI) model trained with human attention data to automatically generate simulated human attention maps as benchmarks for objectively evaluating plausibility of saliency-based XAI methods for image-classification models. Note however that the model involves deep learning with low interpretability. Also, it cannot generate object-specific saliency for object detection models. To our knowledge, no explainable HSI for object detection tasks is currently available to provide benchmarks for plausibility evaluation.

In view of these faithfulness and plausibility issues related to saliency-based XAI methods for computer vision models, here we aimed to examine whether we can embed human attention knowledge into these XAI methods to enhance their faithfulness and plausibility. In image classification or object detection tasks, or more often referred to as object recognition and visual search tasks respectively in the cognitive psychology literature, human participants’ attention strategies as demonstrated in eye movement behavior often reflect their sensitivity to features discriminative of the object class from competitors or diagnostic for efficient target detection and identification (e.g., \cite{RN9, RN13}). Thus, human attention may provide guidance to discriminative or diagnostic features used by the detector that are also accessible and interpretable to humans for XAI methods, potentially enhancing both their faithfulness and plausibility. This guidance may be provided by an interpretable human saliency imitator, which can in turn provide automatically generated human attention maps for benchmarking purposes. 

Accordingly, here we aimed to examine whether human attention can provide guidance to enhance faithfulness and plausibility of saliency-based XAI methods in image classification and object detection. We focused our examinations on two typical gradient-based XAI methods for image classification, Grad-CAM \cite{RN33} and Grad-CAM++ \cite{RN7}. We then developed FullGrad-CAM and FullGrad-CAM++ by extending traditional gradient-based methods to generate explanations for object detection models. We first examined the faithfulness and plausibility of the saliency maps generated by these methods as compared with human attention maps. We then designed a human attention-guided XAI (HAG-XAI) method trained with human attention data that combines explanatory information, such as feature activation maps and their gradients, with an objective to maximize the similarity of the generated XAI saliency maps to the human attention maps. This learnable combination is based on interpretable operations including weighted activation functions and smoothing kernels. We then examined whether the resulting saliency maps have enhanced faithfulness and plausibility in both image classification and object detection, and whether the learned weights could be generalized to another image classification/object detection task/image database to enhance faithfulness and plausibility. Because all learnable parameters in HAG-XAI were interpretable, we thus could understand what led to enhanced faithfulness or plausibility. The HAG-XAI may also be used to generate simulated human attention maps of object detection for benchmarking purposes. 

A preliminary version of our work \cite{RN21} was presented at the 45th Annual Conference of the Cognitive Science Society (CogSci 2023). The major differences of this paper with the preliminary work are: 1) we comprehensively evaluated the HAG-XAI method on different object detection models including both one-stage Yolo-v5s and two-stage Faster-RCNN models; 2) in addition to the object detection task, we collected human attention data in image classification tasks, and tested the HAG-XAI method on popular image classification models with the dataset; 3) the performance of the traditional XAI methods on different layers were tested and compared with the proposed HAG-XAI; 4) more characteristics of the HAG-XAI are discussed in this paper. 
The remainder of this paper is organized as follows: In Study 1, we examined the faithfulness and plausibility of current saliency-based XAI methods and the differences between XAI saliency maps and human attention maps. In Study 2, we introduced the proposed HAG-XAI method and examined its advantages and generalization ability. We then discussed the results and drew conclusions.

\section{Study 1: Comparisons between XAI Saliency Maps and Human Attention Maps}
\subsection{Materials}
\subsubsection{AI Models and Databases}
For image classification, two widely used models were considered, namely the ResNet-50 and Xception. The ImageNet database with one thousand image classes was adopted and the selected models were pretrained on the training set of ImageNet \cite{RN39}. We selected a subset containing 144 images with 18 classes (8 images per class) from the testing set of ImageNet. The selected 18 classes include 8 natural classes (ant, corn, jellyfish, lemon, lion, mushroom, snail, and zebra), and 10 artificial classes (broom, cellphone, fountain, harp, laptop, microphone, pizza, shovel, tennis ball, and umbrella). These image categories were selected since they were from human basic level classes and were also widely used as output classes in image recognition models \cite{RN41, RN42}. Due to the inconsistent image resolution, all images were padded to a unified resolution of 520*400 with white borders. For both models, the input layer were modified to take  input images with a resolution of 520*400. ResNet-50 and Xception achieved an accuracy of 90.97\% and 88.89\% on the selected subset.

For object detection, we focused our examinations on a representative model with a one-stage architecture, Yolo-v5s \cite{RN16}, and a representative model with a two-stage architecture, Faster RCNN \cite{RN30}. In view of the importance of XAI for object detection during automated driving scenarios to ensure safety \cite{RN11}, we selected the BDD-100K \cite{RN37}, a popular well-annotated driving image database, as the target training and evaluation database. All images have a resolution of 1280 x 720. We trained both Yolo-v5s and Faster RCNN using 69400 images from the training set with five types of labels (including ‘car’, ‘truck, ‘bus’, ‘person’, and ‘rider’) from scratch for the whole network using the default training configurations. We then tested the two models with the validation set (containing 10000 images). The two models achieved similar recall performance, with 76.27\%±28.6\% and 75.8\%±29.5\% for faster-RCNN and Yolo-v5s, respectively. We then randomly selected two independent image subsets (test dataset A and B) from the validation set, each containing 160 images, to conduct the experiments, and examine the faithfulness and plausibility of current saliency-based XAI methods on these models and how they were compared with human attention maps. For generalization ability test, the officially released model, which pretrained on MS-COCO database with 80 classes, was employed.

\subsubsection{Human Attention Data}

\begin{figure}[!t]
\centering
\includegraphics[width=2.7in]{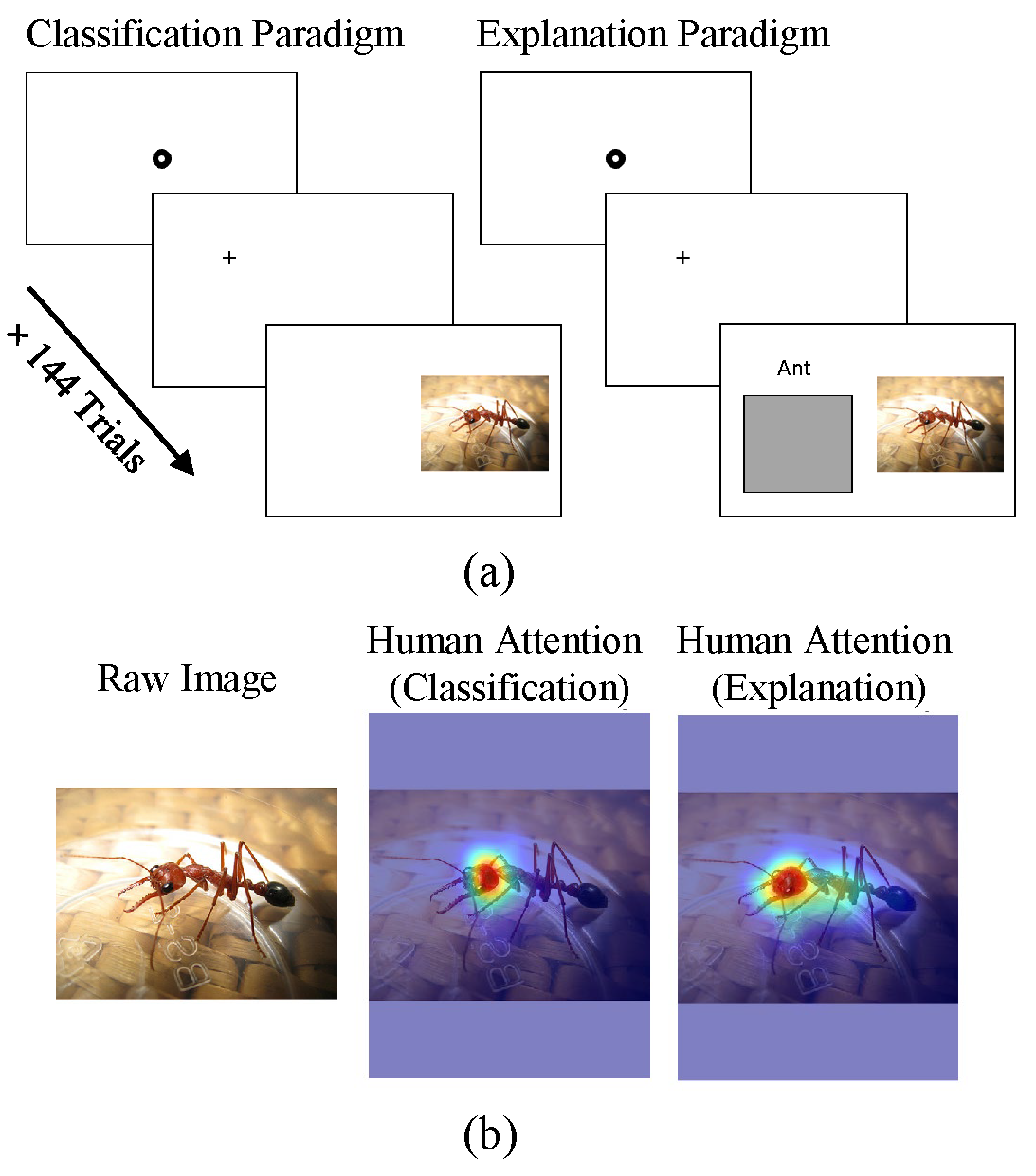}
\caption{Human attention data collection procedure for image classification task. (a) The paradigm for human-based classification task (left) and human-based explanation task (right). (b) Examples of the human attention maps from two different tasks.}
\label{fig_1}
\end{figure}

For image classification, the human attention data were obtained from Qi et al. \cite{RN28}, which included two different tasks: a classification (Cls) task where participants judged the class label of a presented image, and an explanation (Exp) task where participants explained why an image could be classified as a given class label. The study was conducted using a laptop with a screen size of 255 mm × 195 mm and a display resolution of 1024 × 768 pixels. Each image was presented at a size of 9.68° × 12.32° of visual angle when viewed from a distance of 60 cm. The eye fixation data were smoothed by a Gaussian kernel with a standard deviation of 21 pixels (corresponded to 1° visual angle) to generate the human attention map. In the classification task, at the start of each trial, a drift check was performed at the center of the screen. Then, a fixation cross appeared in the upper left corner of the screen. An image was presented on the right side of the screen after the participant fixated on the cross for at least 250 ms, in accordance with the English reading direction (Fig. 1a). After viewing the image, the participant named the class label aloud as quickly as possible, and the image vanished once their response was detected. A similar procedure was used for the explanation task, except that a class label would appear at the location of the initial fixation cross after a stable fixation was detected, and a textbox was available below the label for inputting explanations (Fig. 1a).  Participants were instructed to imagine explaining to someone without prior knowledge, such as a young child. The trial ended when participants hit Enter after completing their explanation. Fig. 1b shows example human attention maps.

\begin{figure}[!t]
\centering
\includegraphics[width=2.8in]{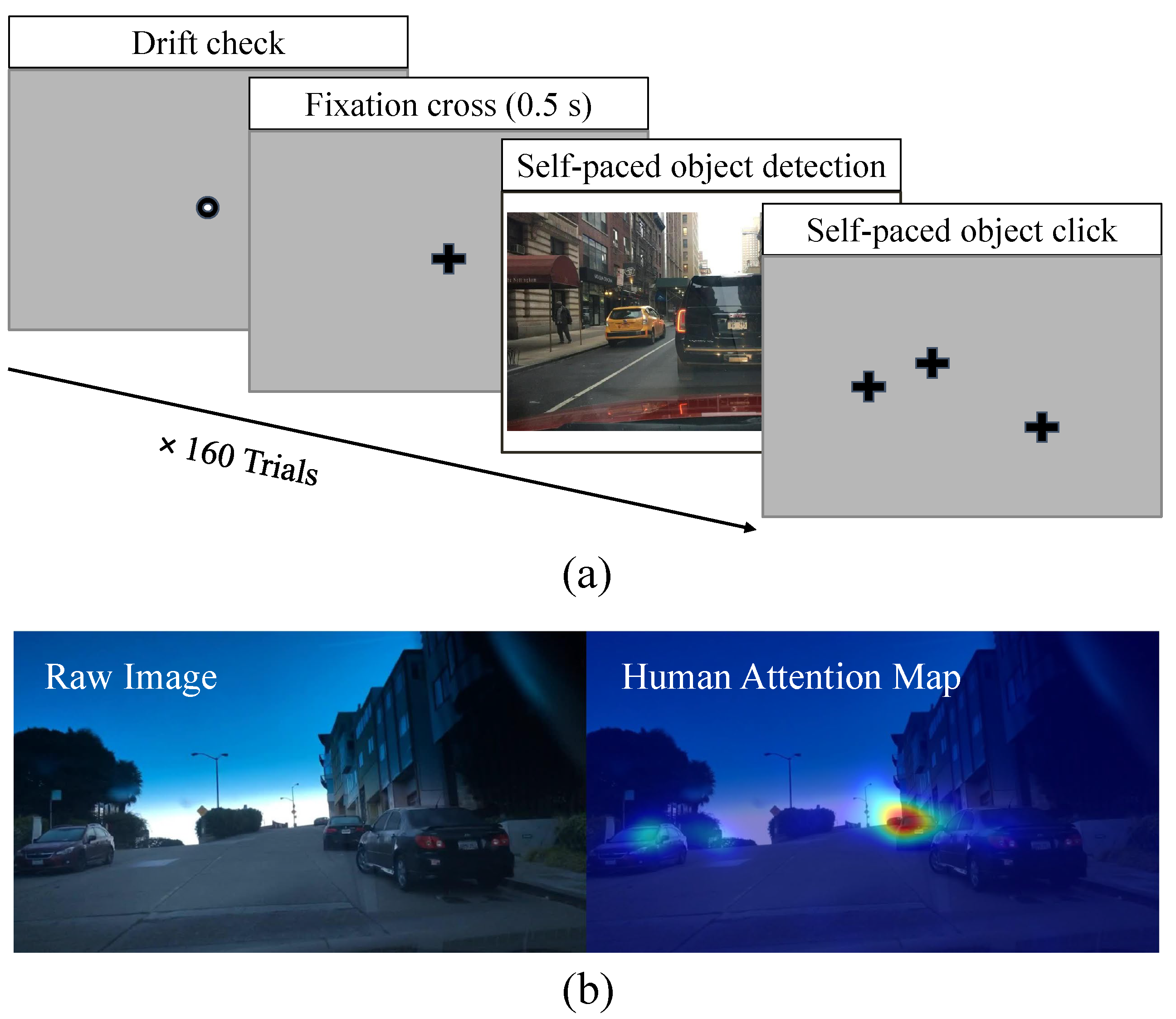}
\caption{Human attention data collection procedure for object detection task. (a) The paradigm for object detection task. (b) An example of the collected human attention map.}
\label{fig_2}
\end{figure}

For object detection, we collected human eye movement data during a vehicle detection task to generate human attention maps. Each trial started with a solid circle at the center of the screen for drift check to ensure stability, followed by a fixation cross for 0.5s. A driving scene image (resized to a resolution of 1024 x 576) was then presented at the center of a 15.6-inch monitor (1920 x 1080-pixel resolution), spanning 34.2° x 20.8° of visual angle under a 55 cm viewing distance. Participants searched for vehicle objects (i.e. ‘car’, ‘truck, and ‘bus’) and pressed the spacebar when they felt they had detected all targets. The screen then turned blank, and participants used a mouse to click on detected target locations (Fig. 2a). We recruited 49 participants to perform the task with images in test dataset A, and 27 participants to perform the task with images in test dataset B. For each subset, eye fixation data of each image over all participants were smoothed by a Gaussian kernel with a standard deviation of 30 pixels, equivalent to one degree of visual angle given the image presentation size. Fig. 2b illustrates an example of the human attention map.

\subsection{Methods}
\subsubsection{XAI Methods}

\begin{figure}[!t]
\centering
\includegraphics[width=2.8in]{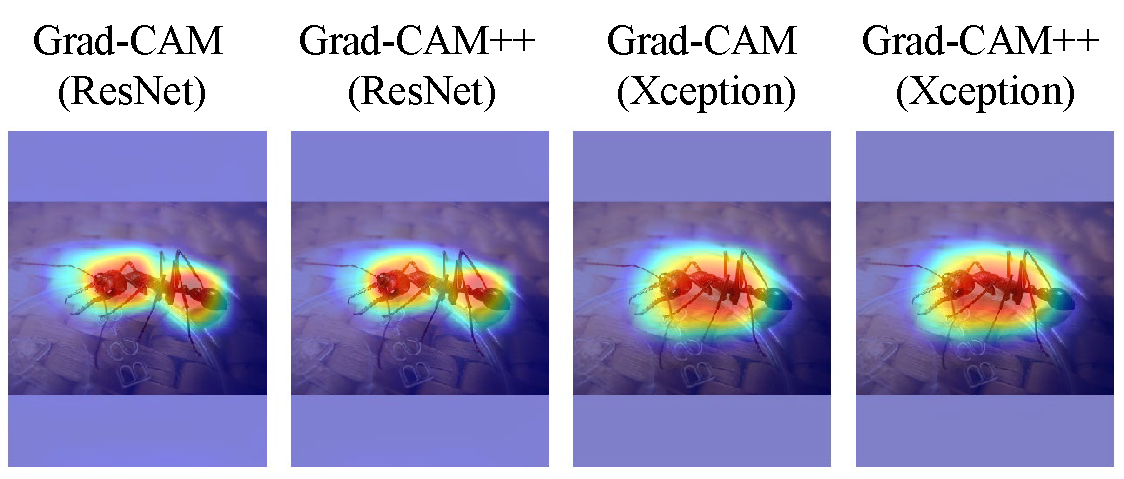}
\caption{Saliency maps generated from different XAI methods and different image classification models.}
\label{fig_3}
\end{figure}

In this study, two typical gradient-based XAI methods: Grad-CAM \cite{RN33} and Grad-CAM++ \cite{RN7}, were utilized for generating explanations for image classification models. Examples of the generated saliency maps for image classification models are illustrated in Fig. 3. The vanilla Grad-CAM method for image classification AI models highlights regions based on the importance of features with respect to a certain class score. However, it did not consider object detection scenarios. Assuming $M(·)$ is an object detection model such as Yolo-v5s, the output of the model with input image $I$ can be expressed as: $y_m = M(I)$, where $y_m$ with $m = 1, 2…, N_{obj}$ is the output classification probability of $m$-th detected object, and $N_{obj}$ is the total number of detected objects. If we apply Grad-CAM to an object detection task, the Grad-CAM map for all detected objects can be expressed as:
\begin{equation}
\label{eq_1}
{S_G} = \sum\limits_{m = 1}^{{N_{obj}}} {\mu \left( {ReLU\left( {\sum\limits_{k = 1}^{{N_{ch}}} {\frac{1}{Z}\sum\limits_{ij} {\frac{{\partial {y^m}}}{{\partial A_{ij}^k}}{A^k}} } } \right)} \right)},
\end{equation}
where $A_k$ is the activation map in the $k$-th layer, $\mu$ is the max-min normalization function that normalizes the data map to scale between 0 to 1, $N_{ch}$ is the number of channels in $A_k$, and ReLU is the rectified linear unit activation function. Additionally, $Z$ is a normalization term defined by a global average pooling operation. As another version of Grad-CAM, Grad-CAM++ modifies the gradient term of vanilla Grad-CAM to improve its interpretability. Accordingly, the Grad-CAM++ for object detection scenarios can be defined as:
\begin{equation}
\label{eq_2}
\begin{array}{l}
S_G^* = \\
\sum\limits_{m = 1}^{{N_{obj}}} {\mu \left( {ReLU\left( {\sum\limits_{k = 1}^{{N_{ch}}} {\frac{1}{Z}\sum\limits_{ij} {\alpha _{ij}^{km}ReLU\left( {\frac{{\partial {y^m}}}{{\partial A_{ij}^k}}} \right){A^k}} } } \right)} \right)},
\end{array}
\end{equation}
where $\alpha_{ij}^{km}$ is a coefficient in $(i, j)$ position for $m$-th detected object to adjust the weight for $k$-th channel of the gradient. A ReLU function is applied to the gradient term to retain the most important features with a positive gradient value.

\begin{figure}[!t]
\centering
\includegraphics[width=2.7in]{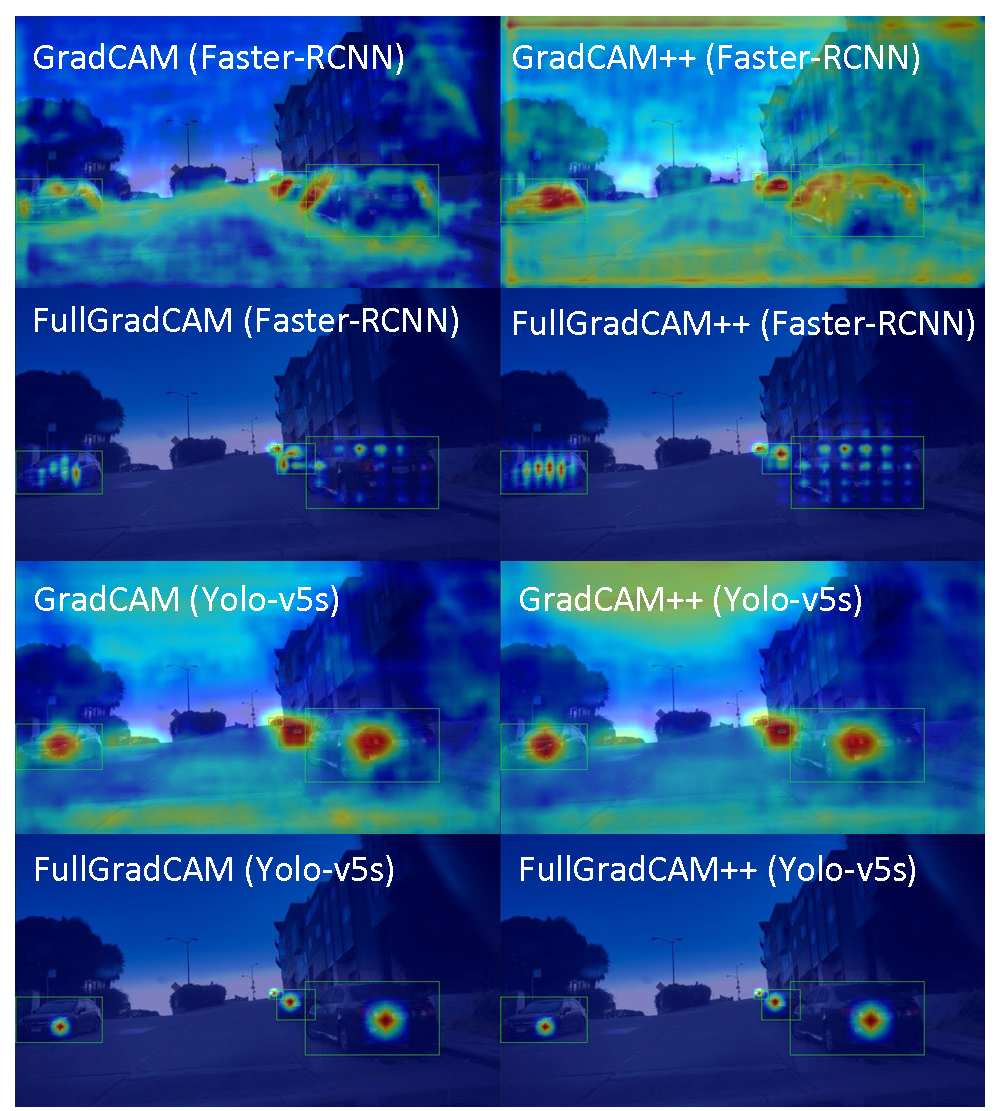}
\caption{Saliency map examples generated using different XAI methods and different object detection models.}
\label{fig_4}
\end{figure}

In object detection models, the gradient maps also contain informative spatial information; however, it was not utilized in vanilla Grad-CAM or Grad-CAM++ due to the global average pooling operation applied to gradients (i.e., summation over $i$ and $j$). As a result, saliency maps generated from Grad-CAM and Grad-CAM++ contain many salient areas not well correlated with detected targets (see Fig. 4). Therefore, here we propose FullGrad-CAM and FullGrad-CAM++ as new XAI methods to generate object-specific saliency maps for object detection models. The FullGrad-CAM method was derived from Grad-CAM, where no average pooling operation is applied to gradients\footnote{Concurrent work by Zhao et al. \cite{RN38} focuses on instance-specific XAI for object detectors, and uses a similar formulation to (3) but does not sum over the objects.  In contrast our work examines the saliency/XAI for all objects in the image, while further comparing to human attention maps for measuring plausibility.}. The FullGrad-CAM is defined as:
\begin{equation}
\label{eq_3}
{S_F} = \sum\limits_{m = 1}^{{N_{obj}}} {\mu \left( {ReLU\left( {\sum\limits_{k = 1}^{{N_{ch}}} {\frac{{\partial {y^m}}}{{\partial {A^k}}} \odot {A^k}} } \right)} \right)},
\end{equation}
where $\odot$ represents the Hadamard product.

When we apply the ReLU function to the gradient term following Grad-CAM++, our FullGrad-CAM++ is defined as:
\begin{equation}
\label{eq_4}
S_F^* = \sum\limits_{m = 1}^{{N_{obj}}} {\mu \left( {ReLU\left( {\sum\limits_{k = 1}^{{N_{ch}}} {ReLU\left( {\frac{{\partial {y^m}}}{{\partial {A^k}}}} \right) \odot {A^k}} } \right)} \right)},
\end{equation}

Previous research \cite{RN33} has suggested that the feature map of the last convolutional layer in deep networks contains the most informative and abstract features. Accordingly, we focused current examinations on saliency maps generated from the last convolutional layer of both image classification and object detection models. For image classification models used in this study, the activated output of the last convolutional layer (also the layer before the global average pool operation) is selected to compute gradient-based saliency maps. As for object detection models, for Yolo-v5s, the last convolutional layer of the whole model was used, while for Faster-RCNN, the last convolutional layer of the backbone module was used in order to obtain global saliency rather than within each detected bounding box. Note that the last convolutional layer of Yolo-v5s belonged to the neck module, which had a multi-scale branch architecture (i.e., small, middle, and large scales). Hence, we first determined which branch each detected object output was from, and then generated the saliency map accordingly. The visual explanations from different XAI methods are shown in Fig. 4. Unlike the visual explanations for image classification models shown in Fig. 3, the saliency maps obtained from traditional gradient-based methods have many noises, which may lead to lower faithfulness and plausibility. Especially, for Faster-RCNN, the resulting saliency maps contained grid-like patterns, which were caused by the application of a $7*7$ ROI pooling operation. For Yolo-v5s, salient areas were more focused due to small activated areas inside the raw gradient term. 

\subsubsection{Faithfulness Evaluation Methods}
We computed the faithfulness using deletion and insertion approaches according to previous studies \cite{RN7, RN26, RN33}. More specifically, the deletion operation deleted salient areas step-by-step according to the saliency scores. The deleted area was filled with random colors. In contrast, the insertion operation inserted salient areas into an empty image with a pure black background step by step according to the saliency scores. For both operations, 100 steps were conducted to record the confidence changes. Particularly, the maximum deletion and insertion area was limited to the summation area of all detected bounding boxes for object detection models. In each step, 1\% of the total area was deleted or inserted, and the deletion or insertion score was computed and recorded (see \cite{RN7}, for details). Finally, a deletion curve or insertion curve reflecting the change of the deletion or insertion score can be obtained. The area under the insertion curve (i-AUC) and deletion curve (d-AUC) were worked as the faithfulness measures.

\subsubsection{Plausibility Evaluation Methods}
Here we used human attention as an objective human-grounded plausibility criterion. Therefore, the plausibility can be defined as the similarity of the XAI saliency maps to human attention maps. Two similarity measures were employed: (1) Pearson Correlation Coefficient (PCC), which could be seen as a relative similarity measure, and is defined as:
\begin{equation}
\label{eq_5}
PCC(u_1, u_2) = \frac{\bar{u}_1^T \bar{u}_2}{{\sqrt{\bar{u}_1^T\bar{u}_1 \bar{u}_2^T\bar{u}_2}}},
\end{equation}
where $u_j$ is a vectorized saliency map, and $\bar{u}_j = u_j - mean(u_j)$ is the mean-subtracted vector of $u_j$. (2) Root Mean Square Error (RMSE), an absolute similarity measure, is defined as:
\begin{equation}
\label{eq_6}
RMSE = \frac{1}{{HW}}{\left\| {{u_1} - {u_2}} \right\|_2}\,
\end{equation}
where $H$ and $W$ are the height and width of the raw image, and $\left\| {·} \right\|_2$ is the $L_2$-norm operator.

\subsection{Results}
\subsubsection{Image classification results}

\begin{figure}[!t]
\centering
\includegraphics[width=3.2in]{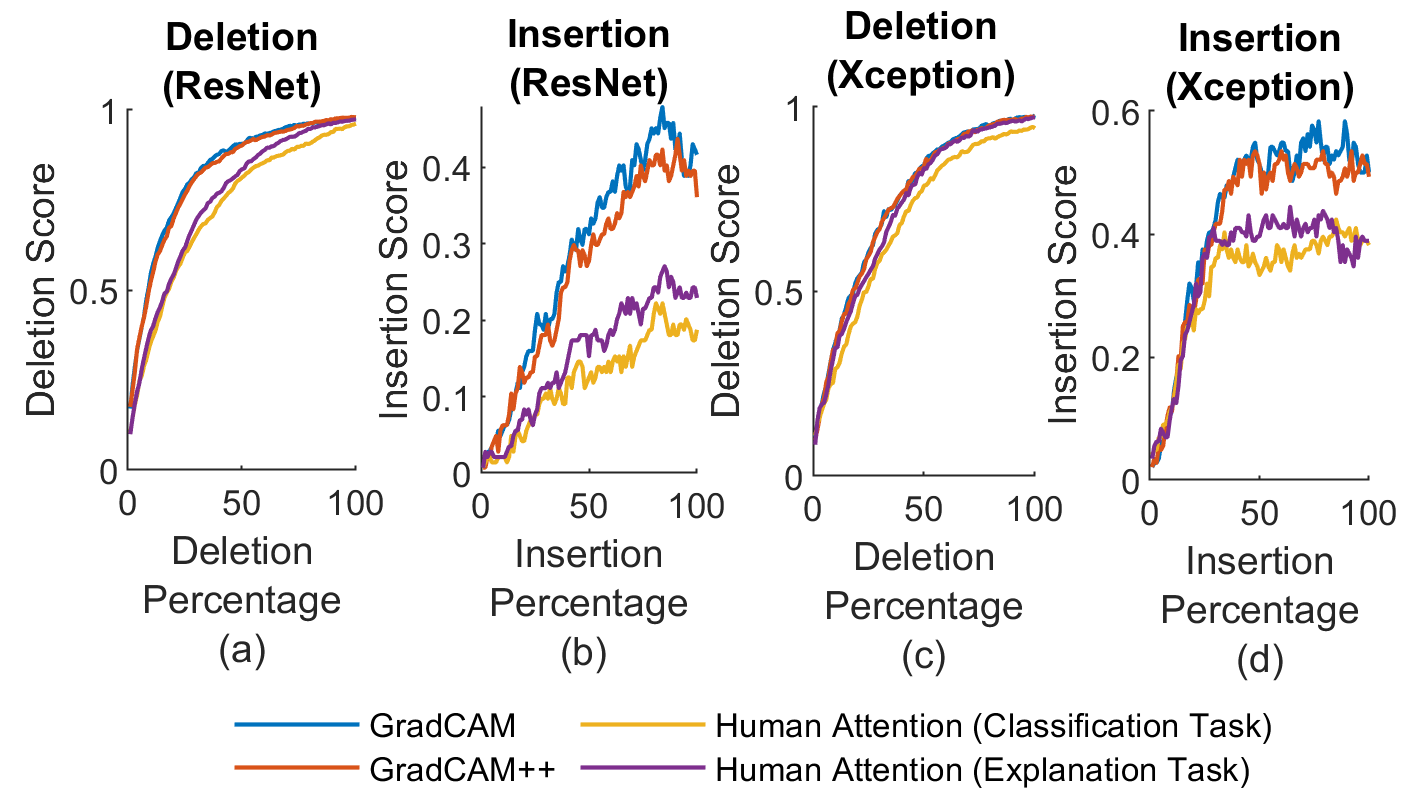}
\caption{Faithfulness (deletion and insertion) of different types of saliency maps for image classification.}
\label{fig_5}
\end{figure}

For image classification, we examined the faithfulness and plausibility of the saliency maps generated using the two Grad-CAM methods for the two image classification models using the test dataset. For faithfulness (Fig. 5), we found that the human attention maps from both the classification and the explanation tasks had lower faithfulness than saliency maps generated from Grad-Cam (gradient-based XAI) methods designed for explaining image classification models. This result suggested that discriminative features humans attend to differed from those used by the models, and thus human attention maps had lower faithfulness than saliency maps generated by XAI methods. Interestingly, human attention maps from the explanation task had better faithfulness than that from the classification task. Previous research has suggested that humans may attend to just sufficient information for making image classification decisions; in contrast, during explanation, they may attend to as much relevant information as possible in order to provide a more comprehensive explanation. Consequently, human attention during explanation may match better with the information used by image classification models than that during the classification task \cite{RN28}. Similarly, human attention during the explanation task had higher similarity to the saliency maps of AI’s attention strategy using the two gradient-based XAI methods, and the two XAI methods achieved similar plausibility performance across the two models (Fig. 6). Since human attention maps had lower faithfulness than the saliency maps generated using current XAI methods, we speculated that embedding human attention knowledge into XAI methods may only enhance their plausibility but not faithfulness. 

\begin{figure}[!t]
\centering
\includegraphics[width=3.2in]{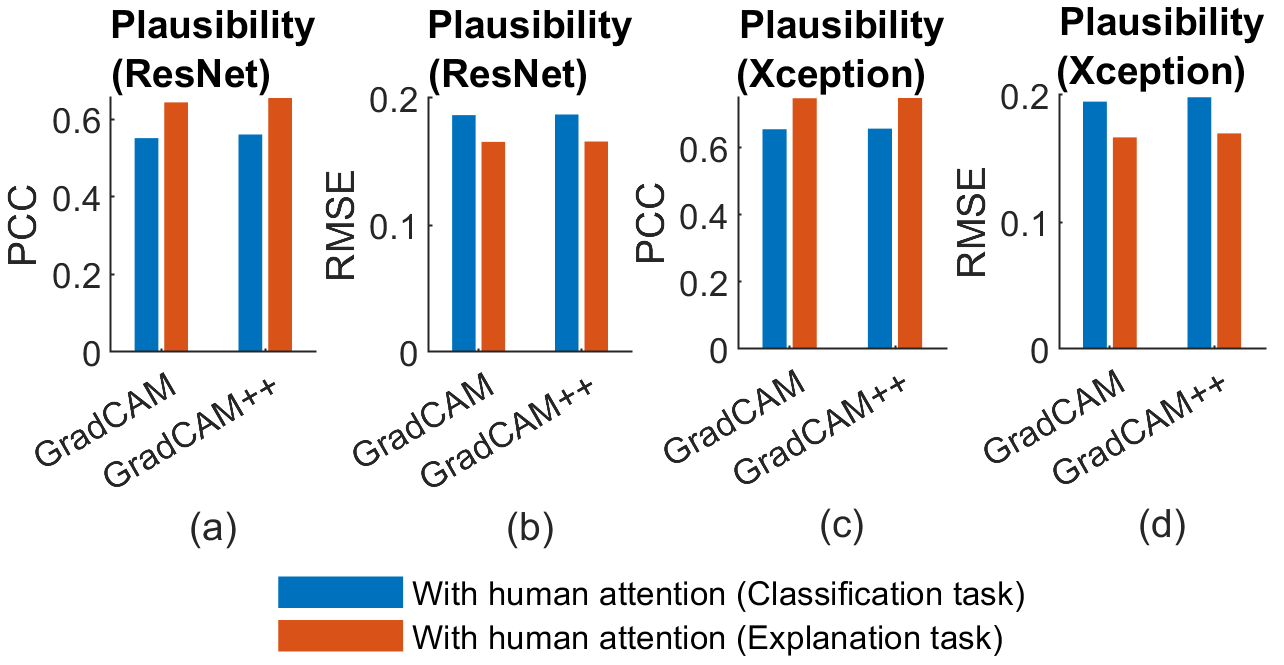}
\caption{Plausibility measures for different image classification models and different XAI methods.}
\label{fig_6}
\end{figure}

\subsubsection{Object detection results}

\begin{figure}[!t]
\centering
\includegraphics[width=3.0in]{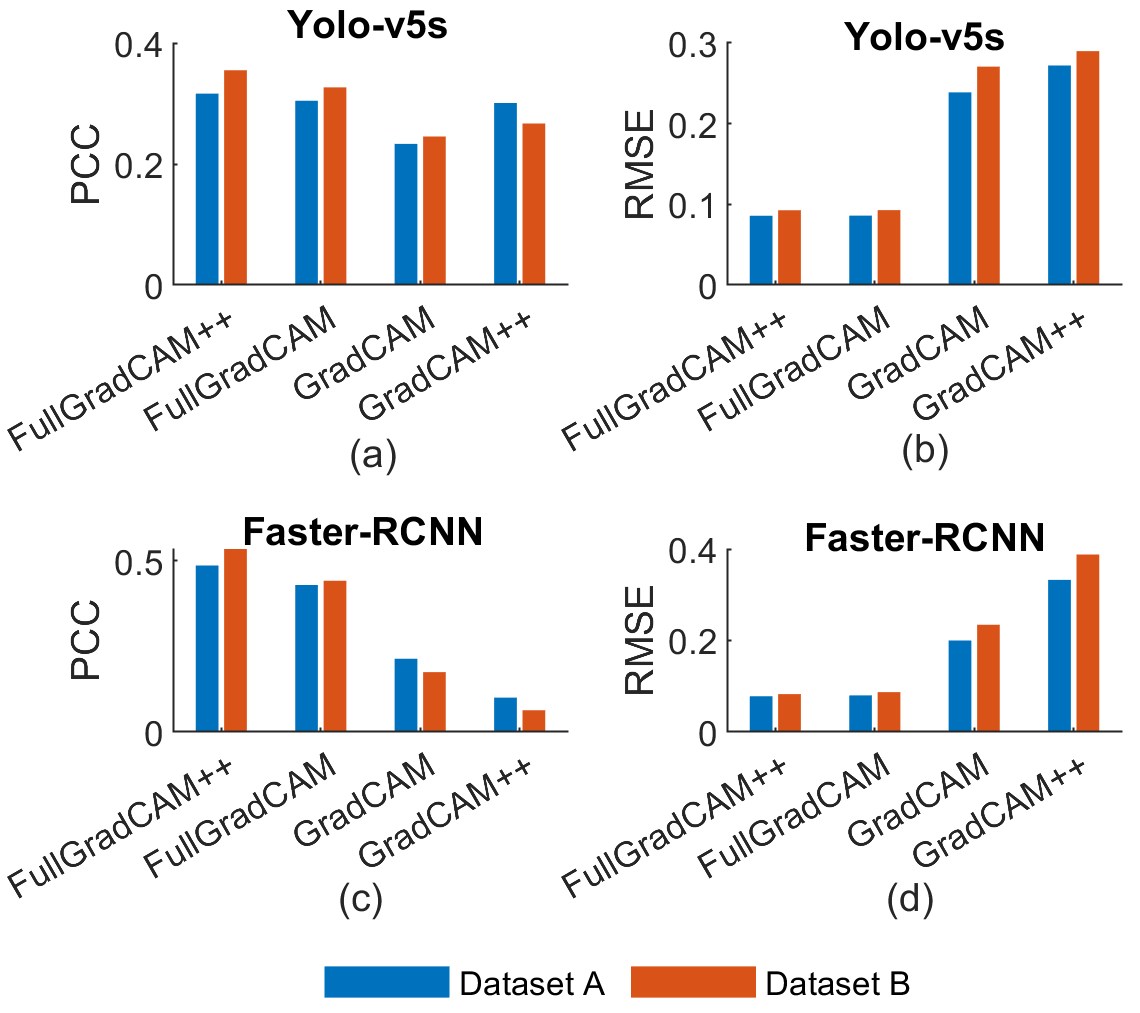}
\caption{Plausibility of XAI saliency maps for Yolo-v5s and Faster-RCNN detection models using the two test datasets. Higher PCC and lower RMSE indicated better plausibility.}
\label{fig_7}
\end{figure}

\begin{figure}[!t]
\centering
\includegraphics[width=3.2in]{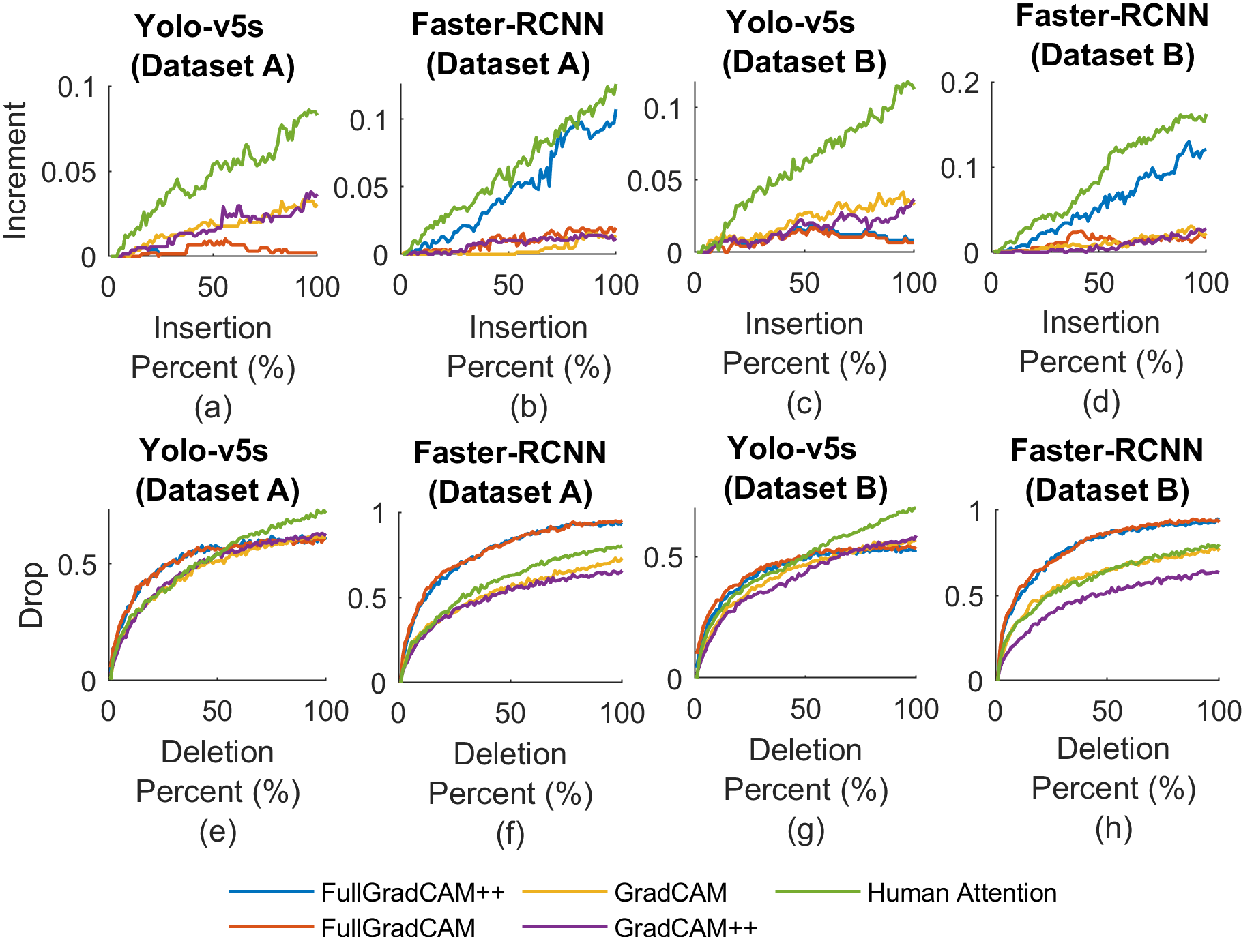}
\caption{Faithfulness of XAI saliency maps vs. human attention maps for Yolo-v5s and Faster-RCNN detection models using the two test datasets. Higher deletion and insertion scores and larger areas under the deletion/insertion curves indicated higher faithfulness. The Y-axis of the plot denotes the increment or drop of the model prediction score.}
\label{fig_8}
\end{figure}

For object detection, we examined the plausibility and faithfulness of the saliency maps generated using the two proposed FullGrad-CAM methods for the two object detection models using the two test datasets. In plausibility as measured in PCC and RMSE (Fig. 7), the plausibility of the proposed FullGrad-CAM and FullGrad-CAM++ methods were consistently better than Grad-CAM and Grad-CAM++ methods on the two object detection models and two test datasets. In addition, FullGrad-CAM++ achieved better plausibility than FullGrad-CAM, suggesting that human attention during the task may mainly focus on features that lead to increased object class scores. In faithfulness using the insertion approach (Fig. 8), interestingly, human attention map had higher faithfulness in most cases when being used as an XAI saliency map as compared with those from existing saliency-based XAI methods, although our participants had no knowledge of the operations of the two AI models. Similar findings were observed using the deletion approach for Yolo-v5s. For Faster-RCNN, although FullGradCam and FullGradCam++ seemed to outperform human attention maps in faithfulness using the deletion approach, previous research has suggested that the fragmented saliency areas generated by Faster-RCNN may seriously affect the deletion score, and thus the deletion score may not accurately reflect faithfulness of these XAI methods on Faster-RCNN \cite{RN6, RN10, RN15}.  Together our results suggested that current saliency-based XAI methods for object detection AI models did not well capture meaningful features used by the models, with their faithfulness scores lower than attention maps generated by human participants performing the same task but with no knowledge of the models’ operations. This result also suggested that these object detection AI models may be using similar information extraction strategies to humans, resulting in high faithfulness when using human attention maps as XAI saliency maps. This finding also justified the use of human attention maps as benchmarks for current saliency-based XAI methods.  

When we compared the faithfulness scores of FullGrad-CAM and FullGrad-CAM++, the ReLU function applied to the gradient term in FullGrad-Cam++ enhanced its faithfulness on the faster-RCNN model. FullGrad-CAM++ also achieved higher plausibility. Accordingly, we speculated that operations on gradients could significantly affect faithfulness and plausibility of XAI methods. In addition, the comparison between XAI-generated saliency maps and human attention maps suggested that smoothing strategies on both gradients and activation may also affect faithfulness. This motivated us to develop an explainable method to find the optimal activation functions and smoothing strategies for enhancing XAI saliency maps’ faithfulness and plausibility using human attention data, in Study 2 (Section III).

Based on the result that human attention maps had better faithfulness than saliency maps from traditional XAI methods for explaining object detection models, we further investigate image-level correlations between the faithfulness and plausibility of FullGrad-CAM-based XAI methods. The correlation coefficient and its significance for all pairs of faithfulness (deletion and insertion scores) and plausibility (similarity measured by PCC) are computed. In total, 2 (AI models) × 2 (Faithfulness including Deletion and Insertion) = 4 correlation results were obtained, shown in Table I. Note that higher PCC indicates higher plausibility. In general,  higher plausibility (similarity to human attention) is associated with higher faithfulness when using the insertion method. It can be speculated from the results that humans and AI may have similar attention strategies, but current XAI methods cannot well describe these strategies. This further motivates us to use the human attention map as the guidance to improve the faithfulness of XAI particularly for object detection models.

\begin{table}[]
\centering
\caption{Image-level correlation results between faithfulness and similarity}
\label{tab:Table_1}
\resizebox{\columnwidth}{!}{%
\begin{tabular}{ccccccc}
\hline
\textbf{No.} & \textbf{Model} & \textbf{XAI} & \textbf{Faithfulness} & \textbf{Similarity} & \textit{\textbf{r}} & \textit{\textbf{p}} \\ \hline
1 & RCNN & FullGrad-CAM-based & Insertion & PCC & 0.0900 & 0.023 \\
2 & RCNN & FullGrad-CAM-based & Deletion & PCC & -0.0441 & 0.275 \\
3 & YOLO & FullGrad-CAM-based & Insertion & PCC & 0.1775 & \textless{}0.001 \\
4 & YOLO & FullGrad-CAM-based & Deletion & PCC & -0.0256 & 0.525 \\ \hline
\multicolumn{7}{l}{\textit{Note: FullGrad-CAM-based methods include FullGrad-CAM and FullGrad-CAM++.}}
\end{tabular}%
}
\end{table}

\section{Study 2: Human Attention-Guided XAI}
\subsection{Methods}
In this study, we aim to leverage human attention information  to guide the combination of explanatory information, including activation maps and gradient maps, for gradient-based XAI methods in computer vision models, which can potentially enhance their plausibility and faithfulness. We proposed Human Attention Guided XAI (HAG-XAI), which has learnable activation functions and smoothing kernels for both the gradient and activation terms. The input of the model is the raw activation map (feature map) and gradient tensors. Since values in the gradient and activation maps might have arbitrary scales (due to the multiplicative nature of the layers in the model), we reweight them to obtain a better combination with optimal plausibility. The smoothing kernels are used to accommodate differences in local feature aggregation and receptive field sizes between CNNs and humans. Although HAG-XAI has trainable parameters, its operations are simple and interpretable.

\subsubsection{Learnable activations and smoothing kernels}
In more detail, different from the fixed weight scales for gradient and activation maps used in traditional gradient-based XAI methods, we provide an adaptive piecewise linear activation function with two learnable parameters:
\begin{equation}
\label{eq_7}
\varphi _{{\alpha ^ - }}^{{\alpha ^ + }}\left( \theta  \right) = {\alpha ^ + }\max \left( {\theta ,0} \right) + {\alpha ^ - }\min \left( {\theta ,0} \right),
\end{equation}
where $\theta$ is the input activation map. The two learnable parameters $\alpha^+$ and $\alpha^-$ allow for different scalings (or complete truncation) of the positive and negative parts of the activation, respectively. For example, the standard ReLU activation is obtained when $\alpha^+=1$ and $\alpha^-=0$, and the absolute value function is obtained when $\alpha^+=1$ and $\alpha^-=-1$, and the linear activation is obtained when $\alpha^+=1$ and $\alpha^-=1$. Thus the training of the learnable activation function will both select and weigh the important positive or negative components of the input map. During training, the two parameters used for the activation map ($\alpha^+$ and $\alpha^-$) are initialized to 1 (equivalent to a linear activation function), while the two parameters for the gradient map (denoted as $\beta^+$ and $\beta^-$) are initialized to 1 and 0 (equivalent to ReLU function). 
Smoothing kernels are applied to the gradient map and final saliency map to better highlight neighboring features. The gradient can be aggregated over local regions by smoothing to enhance the plausibility. Meanwhile, adding a smoothing operation to the whole saliency map models the difference between the receptive field sizes of the human and the AI model, as well as the sensor noise of the eye tracker used to measure human attention The smoothing is implemented with a learnable 2D Gaussian kernel of size 21 × 21 for object detection models, and of size 9 × 9 for image classification models. The learnable Gaussian smoothing kernel is defined as:
\begin{equation}
\label{eq_8}
G_A^v\left( {x,y} \right) = A\exp \left( { - \frac{{{{\left( {x - {x_c}} \right)}^2} + {{\left( {y - {y_c}} \right)}^2}}}{{2\left| v \right| + \varepsilon }}} \right),
\end{equation}
where $(x,y)$ is the spatial coordinate, $(x_c,y_c)$ is the constant mean set to the half length of the kernel size, $v$ is the learnable variance with an initialization value of 3 for object detection models, and of 1 for image classification models. $A$ is the learnable amplitude with an initialization value of 1, and $\epsilon$ is a small constant to avoid dividing by zero.

\subsubsection{HAG-XAI}
Our HAG-XAI saliency generation method is:
\begin{equation}
\label{eq_9}
\begin{array}{l}
{S_{HI}} = G_{{A_\gamma }}^{{v_\gamma }} * \\
\sum\limits_{m = 1}^{{N_{obj}}} {\bar \mu \left( {ReLU\left( {\sum\limits_{k = 1}^{{N_{ch}}} {\left( {G_{{A_\alpha }}^{{v_\alpha }} * \varphi _{{\alpha ^ - }}^{{\alpha ^ + }}\left( {\frac{{\partial {y^m}}}{{\partial {A^k}}}} \right)} \right) \odot \varphi _{{\beta ^ - }}^{{\beta ^ + }}\left( {{A^k}} \right)} } \right)} \right)},
\end{array}
\end{equation}
where $\varphi _{{\alpha ^ - }}^{{\alpha ^ + }}$ and $\varphi _{{\beta ^ - }}^{{\beta ^ + }}$ are the learnable activations for the gradient and activation map, $G_{{A_\gamma }}^{{v_\gamma }}$ and $G_{{A_\alpha }}^{{v_\alpha }}$ are learnable Gaussian smoothing kernels for the final map and the gradient map, and $*$ is the convolution operator. Note that the same kernel is applied to each channel of gradient and activation tensors. Meanwhile, $\bar{\mu}$ is an object normalization function for object detection models. During visual search, human participants tended to attend to small objects more than large objects. Accordingly, we normalize each object’s individual saliency map according to their activated area:
\begin{equation}
\label{eq_10}
\bar \mu \left( \theta  \right) = \frac{\theta }{{\sum\limits_{ij} {\theta  + \varepsilon } }},
\end{equation}
where $\epsilon$ is a small constant value to avoid the denominator being 0. In total, only eight learnable parameters need to be optimized in the training stage of the HAG-XAI method. For image classification tasks, the HAG-XAI method does not include the normalization function, and Nobj is equivalent to 1 for each image.
The training goal of this model is to obtain a human-like saliency map. Therefore, the loss function is set to the (dis)similarity between human attention map and AI saliency map, based on PCC and RMSE. The optimization objective of the model is:
\begin{equation}
\label{eq_11}
\mathop {\arg \min }\limits_\theta  \left\{ \begin{array}{l}
1 - \frac{{{{\left( {S_{HI}^*} \right)}^T}S_H^*}}{{\sqrt {{{\left( {S_{HI}^*} \right)}^T}S_{HI}^*{{\left( {S_H^*} \right)}^T}S_H^*} }}\\
 + \frac{1}{{HW}}{\left( {{{\left\| {S_{HI}^* - S_H^*} \right\|}_2}} \right)^2}
\end{array} \right\},
\end{equation}
where $S_{HI}^*$ and $S_H^*$ are the flattened XAI saliency map and human attention map (serving as ground-truth) in the training set, $\theta$ is the parameters of the whole model.

\subsubsection{Experiment setup}
To test our method, for image classification, we conducted a five-fold cross-validation on the collected ImageNet subset and tested the resulting saliency maps’ faithfulness and plausibility. Results were reported with the averaged performance of five-fold validation data. For object detection, we used the BDD-100K database used in Study 1 to train the HAG-XAI and tested their faithfulness and plausibility. During training, test dataset A from Study 1 was used as the training/validation set and dataset B as the testing set. The training set was divided into five parts to conduct a five-fold cross-validation for learning the HAG-XAI parameters. The well-trained HAG-XAI models were assessed on both validation set of dataset A and testing set (dataset B) after training. On the dataset A, the averaged 5-fold cross-validation performance was reported. Five models generated from the 5-fold cross-validation procedure were assessed on the whole testing set, and the averaged testing performance was reported. Considering the neck module of Yolo-v5s has three different scales, the activations and gradients were resized to a uniform (maximum) spatial resolution before training. For all tasks, the Adam optimizer was used during training, with the minibatch size set to 30 and 144, and the maximum number of training epochs set to 120 and 200 for object detection and image classification tasks, respectively. The learning rate was set to 0.05 initially and exponentially decreased to 0.005 within 120 epochs. An early stop strategy on validation set was employed in the training procedure of object detection HAG-XAI model to boost the training speed and avoid overfitting, where the patience was set to 30 epochs. For image classification HAG-XAI model, no early stop strategy was performed since no overfitting problem was observed in training stage after our checking.

Additionally, for cases where HAG-XAI methods led to enhanced faithfulness, we evaluated the generalization ability of the learned parameters in HAG-XAI. For object detection, we used the validation set of the MS-COCO object detection database, which contained 5000 images with 80 general object classes \cite{RN20}.

\subsection{Results}

\begin{figure}[!t]
\centering
\includegraphics[width=3.2in]{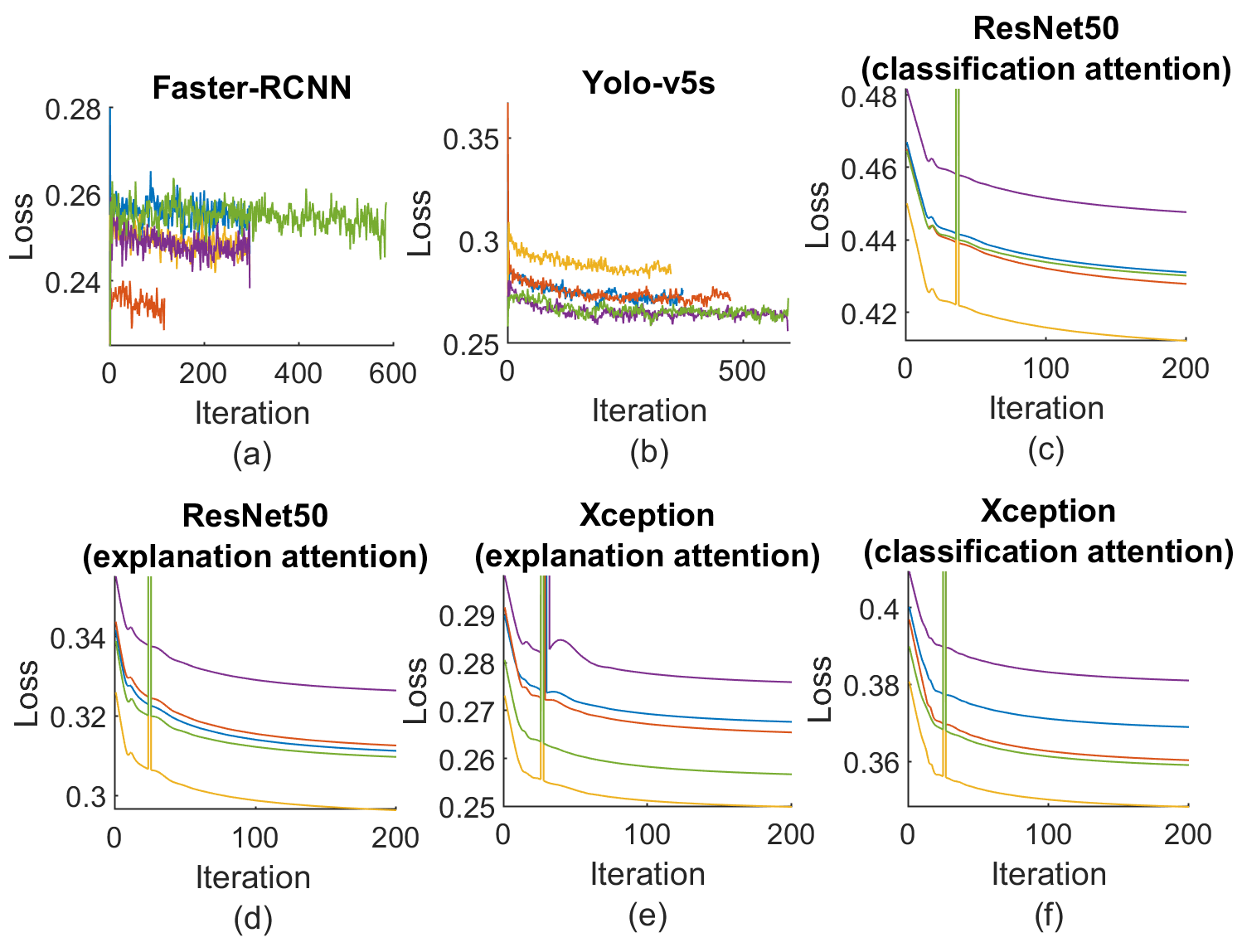}
\caption{The training loss during HAG-XAI training. (a) and (b) are for the object detection models, while (c) to (f) are for the image classification models. Different colors indicate loss curves from different validation folds. Note that an unexpected loss value occured at about 40-th iteration due to outlier data for image classification; it did not affect the convergence of the training procedure.}
\label{fig_9}
\end{figure}

\begin{figure}[!t]
\centering
\includegraphics[width=3.2in]{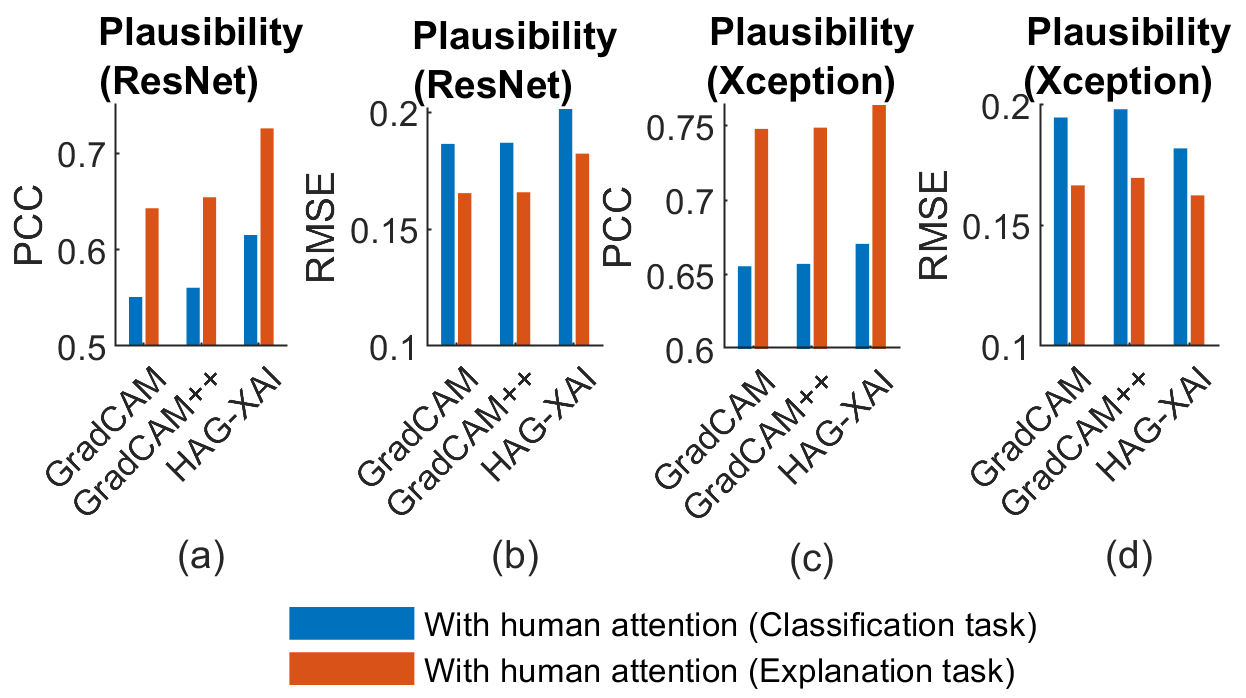}
\caption{The plausibility measures obtained from HAG-XAI and other gradient-based XAI methods for image classification.}
\label{fig_10}
\end{figure}

\begin{figure}[!t]
\centering
\includegraphics[width=2.8in]{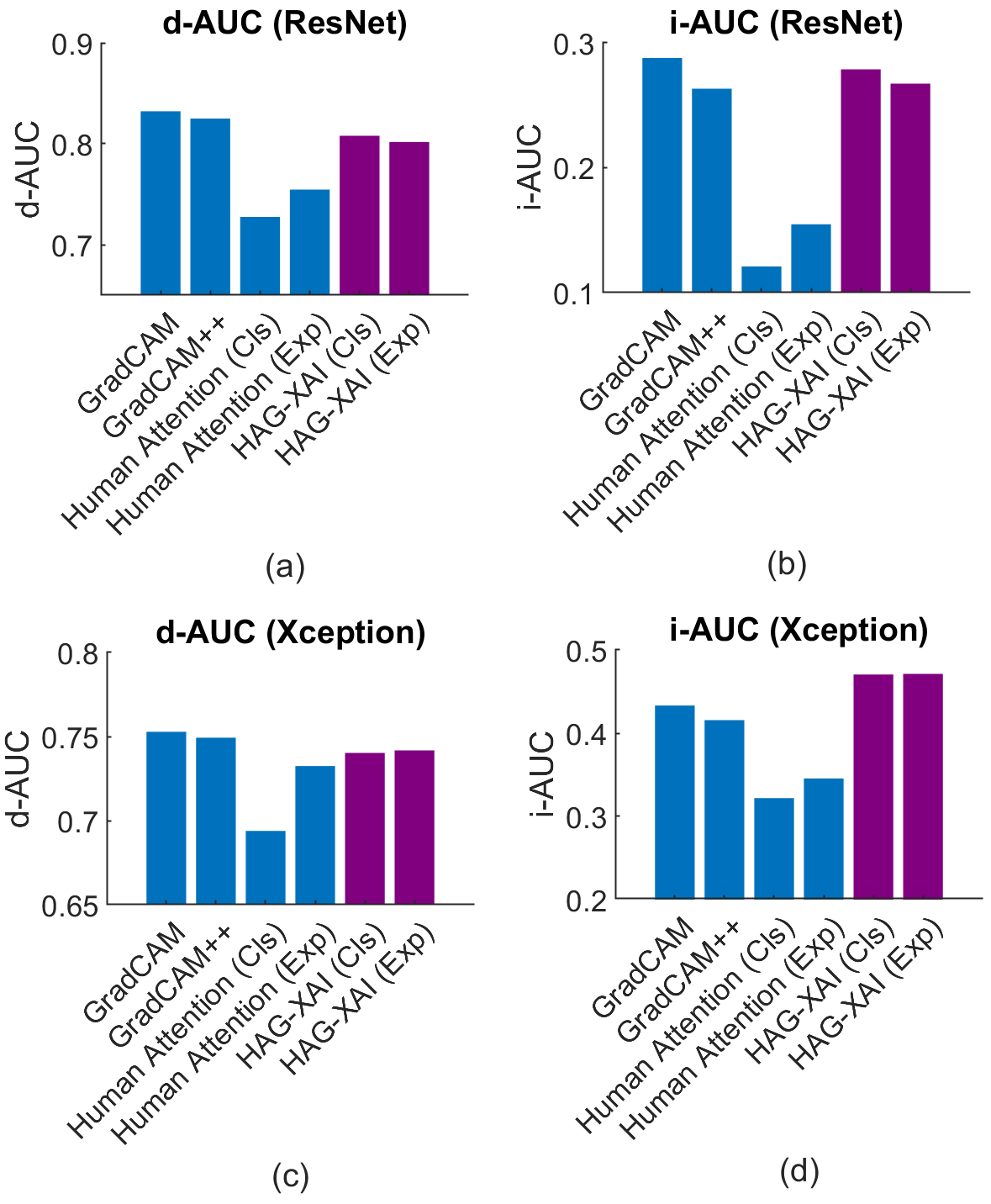}
\caption{The faithfulness measures obtained from HAG-XAI, human attention, and other gradient-based XAI methods.}
\label{fig_11}
\end{figure}

The training loss of HAG-XAI for different tasks and models  converged in about 100 iterations (Fig. 9). For image classification, HAG-XAI achieved a better plausibility in most cases than other XAI methods (Fig. 10). The plausibility as measured in PCC was significantly improved in all cases after applying HAG-XAI, with a high PCC of over 0.7 obtained using human attention data from the explanation task. Saliency maps from Xception had higher plausibility than ResNet. Xception is designed with depth-wise separable convolutions, which are known to capture spatial information more efficiently than traditional convolutional layers \cite{RN40}. This could make it easier for the model to attend to important image regions. In contrast, HAG-XAI resulted in slightly lower faithfulness than other XAI methods in most cases, and only the i-AUC obtained from Xception was higher than other methods (Fig. 11). This result suggested that HAG-XAI may enhance the plausibility at the expense of faithfulness.

\begin{figure}[!t]
\centering
\includegraphics[width=2.8in]{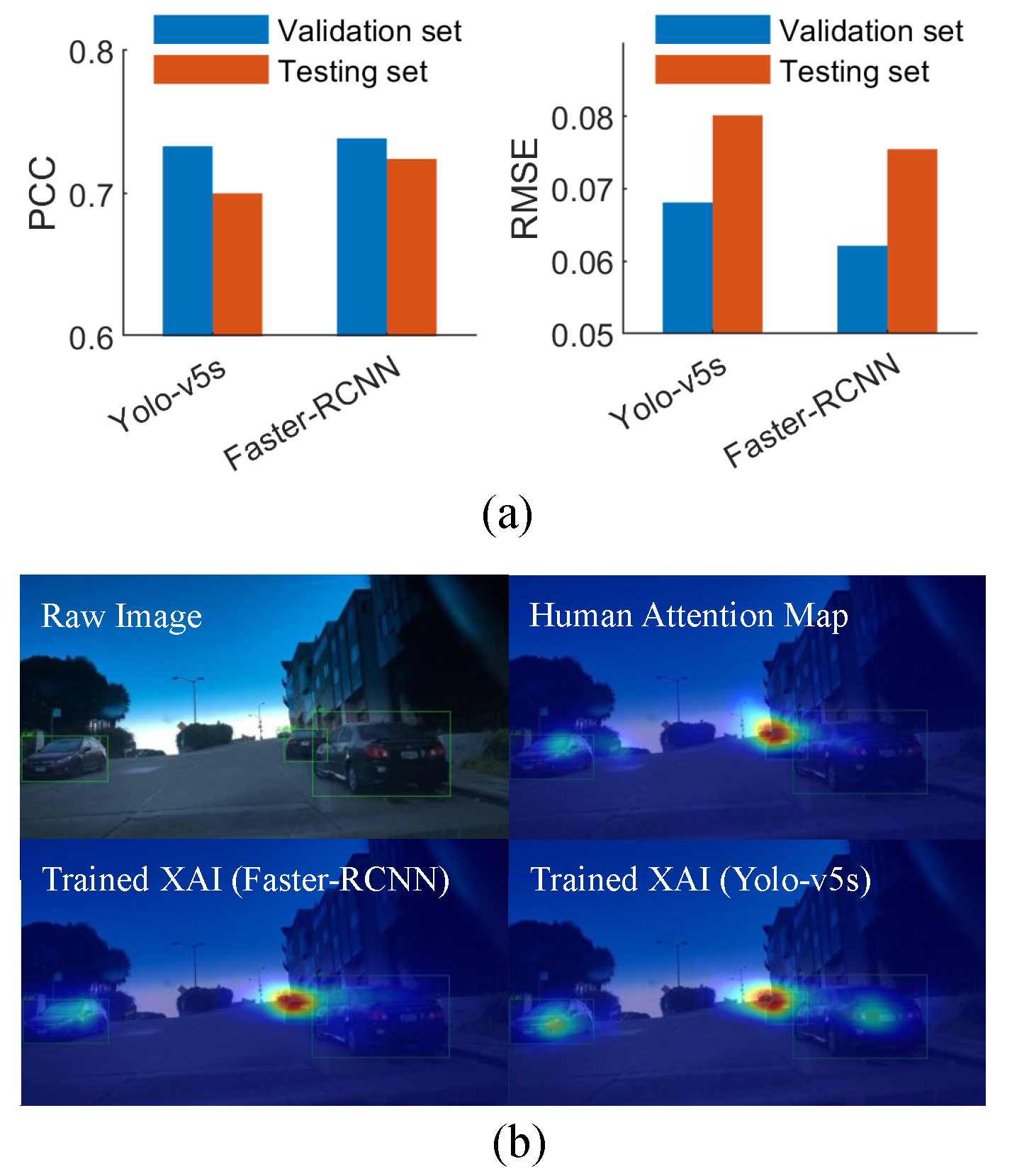}
\caption{(a) Similarity between the trained HAG-XAI saliency maps and human attention maps using different datasets and different object detection models. (b) Example HAG-XAI saliency maps vs. human attention maps.}
\label{fig_12}
\end{figure}

\begin{figure}[!t]
\centering
\includegraphics[width=2.5in]{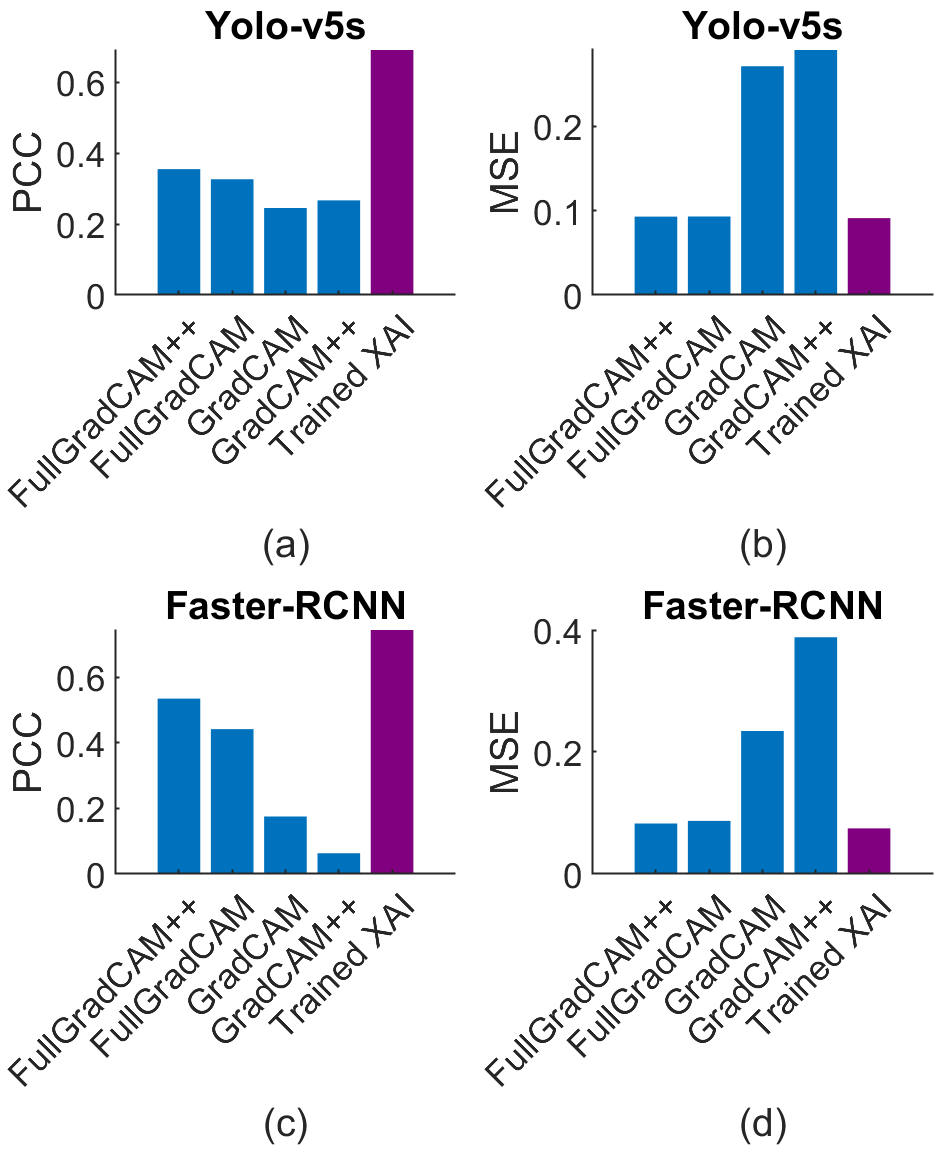}
\caption{The similarity between XAI saliency maps and human attention maps (plausibility) across different XAI methods using different object detection models.}
\label{fig_13}
\end{figure}

\begin{figure}[!t]
\centering
\includegraphics[width=3.2in]{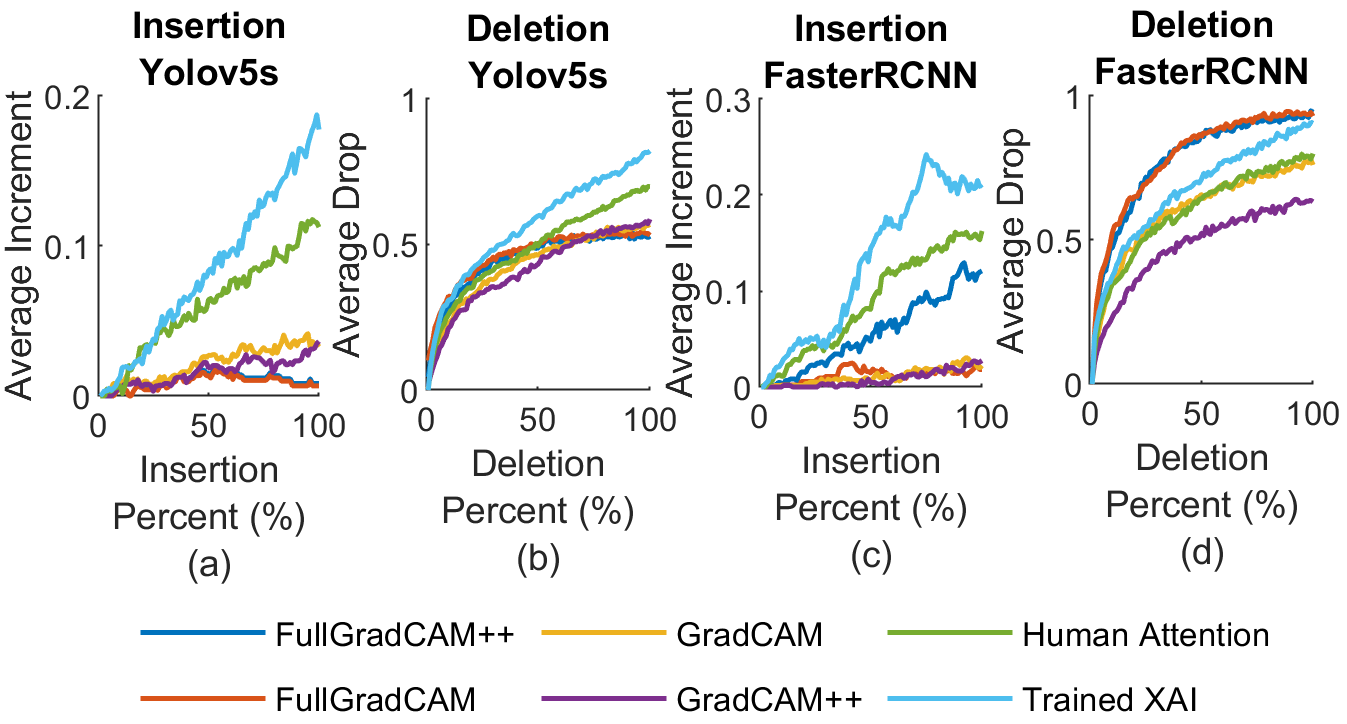}
\caption{The faithfulness performance comparison between different XAI methods and different object detection models.}
\label{fig_14}
\end{figure}

For object detection,  the saliency maps generated from HAI-XAI and human attention maps had high similarity (above 0.7 in PCC on the testing set; Fig. 12b), and was higher than other XAI methods (Fig. 13), indicating a high plausibility. The HAG-XAI also had higher faithfulness than other XAI methods (Fig. 14; except for the deletion score for Faster-RCNN, which could be seriously affected by the fragmented saliency areas). These results suggest that human attention could be used to guide saliency-based XAI to enhance both of its faithfulness and plausibility for object detection applications, in contrast to image classification applications. Compared with untrained saliency maps (Fig. 4), the trained saliency maps (Fig. 12b) were more similar to the human attention maps for both Faster-RCNN and Yolo-v5s models. Finally, the performance difference between the validation and testing sets was small, demonstrating a great generalization ability (Fig. 12a).

We then used images from MS-COCO database to examine whether the learned combination from HAG-XAI could be transferred to other object detection tasks of the same models. Since the image resolutions in MS-COCO were different from BDD-100K, the resolutions of the activation and gradient maps from MS-COCO were adjusted to be the equivalent size of BDD-100K. Tables II and III illustrate the area under the insertion curve (i-AUC) and deletion curve (d-AUC) for assessing faithfulness of the generated saliency maps over 5000 images for detecting the most common categories in the database. The results showed that our method achieved the best deletion and insertion scores across different object detection tasks, demonstrating a great generalization ability of HAG-XAI to other databases/object detection tasks.

\begin{table}[]
\centering
\caption{The d-AUC score when using the learned functions from HAG-XAI to generate saliency maps for object detection tasks in MS-COCO. ‘All’ class refers to the case when all bounding boxes regardless of class were included to compute the faithfulness. ‘Random’ class refers to the case when we randomly selected one object from each image to evaluate its object-specific explanation ability.}
\label{tab:Table_2}
\resizebox{\columnwidth}{!}{%
\begin{tabular}{cccccc}
\hline
\textbf{Classes} & \textbf{FGC*} & \textbf{FGC} & \textbf{GC*} & \textbf{GC} & \textbf{HAG-XAI} \\ \hline
All & 0.521 & 0.521 & 0.729 & 0.741 & \textbf{0.756} \\
Person & 0.380 & 0.382 & 0.640 & 0.651 & \textbf{0.683} \\
Car & 0.716 & 0.718 & 0.726 & 0.720 & \textbf{0.737} \\
Chair & 0.647 & 0.647 & 0.728 & 0.724 & \textbf{0.832} \\
Book & 0.770 & 0.770 & 0.706 & 0.804 & \textbf{0.908} \\
Bottle & 0.704 & 0.703 & 0.818 & 0.859 & \textbf{0.892} \\
Random & 0.407 & 0.407 & 0.633 & 0.647 & \textbf{0.679} \\ \hline
\multicolumn{6}{l}{\textit{\begin{tabular}[c]{@{}l@{}}Note: FGC*: FullGrad-CAM++. FGC: FullGrad-CAM. \\ GC*: Grad-CAM++. GC: Grad-CAM.\end{tabular}}}
\end{tabular}%
}
\end{table}

\begin{table}[]
\centering
\caption{The i-AUC score on MS-COCO}
\label{tab:Table_3}
\resizebox{\columnwidth}{!}{%
\begin{tabular}{cccccc}
\hline
\textbf{Classes} & \textbf{FGC*} & \textbf{FGC} & \textbf{GC*} & \textbf{GC} & \textbf{HAG-XAI} \\ \hline
All & 0.004 & 0.004 & 0.103 & 0.109 & \textbf{0.133} \\
Person & 0.006 & 0.006 & 0.086 & 0.101 & \textbf{0.146} \\
Car & 0.019 & 0.019 & 0.082 & 0.086 & \textbf{0.105} \\
Chair & 0.001 & 0.001 & 0.028 & 0.033 & \textbf{0.058} \\
Book & 0.000 & 0.000 & 0.023 & 0.030 & \textbf{0.033} \\
Bottle & 0.004 & 0.004 & 0.090 & 0.107 & \textbf{0.128} \\
Random & 0.005 & 0.005 & 0.122 & 0.129 & \textbf{0.160} \\ \hline
\multicolumn{6}{l}{\textit{\begin{tabular}[c]{@{}l@{}}Note: FGC*: FullGrad-CAM++. FGC: FullGrad-CAM. \\ GC*: Grad-CAM++. GC: Grad-CAM.\end{tabular}}}
\end{tabular}%
}
\end{table}

\section{Discussion}
Here we examined whether we can embed human attention knowledge into saliency-based XAI methods for computer vision models, including image classification and object detection models, to enhance their faithfulness and plausibility. We found that for image classification models, embedding human attention knowledge generally enhanced plausibility of the saliency maps from current XAI methods at the expense of faithfulness. Given that these XAI saliency maps already have higher faithfulness than human attention maps, this result suggests that the use of some features in humans differs from that in AI models in image classification, and thus embedding human attention information harms faithfulness of the explanations. In contrast, surprisingly for object detection models, human attention maps have higher faithfulness for explaining AI models’ feature use than the saliency maps from the current XAI methods, and consequently embedding human attention knowledge simultaneously enhanced both plausibility and faithfulness. This result suggested that in object detection, humans and AI models may use similar diagnostic features; however, current saliency-based XAI methods did not well capture meaningful features used by the models. The issue may be in the combination of scale weights and aggregation of features across different gradient and activation maps that can faithfully reflect the importance of the features, and human attention helps guide the search for a better combination to enhance the faithfulness.

\subsection{Interpretation of learned HAG-XAI parameters}

\begin{figure}[!t]
\centering
\includegraphics[width=3.2in]{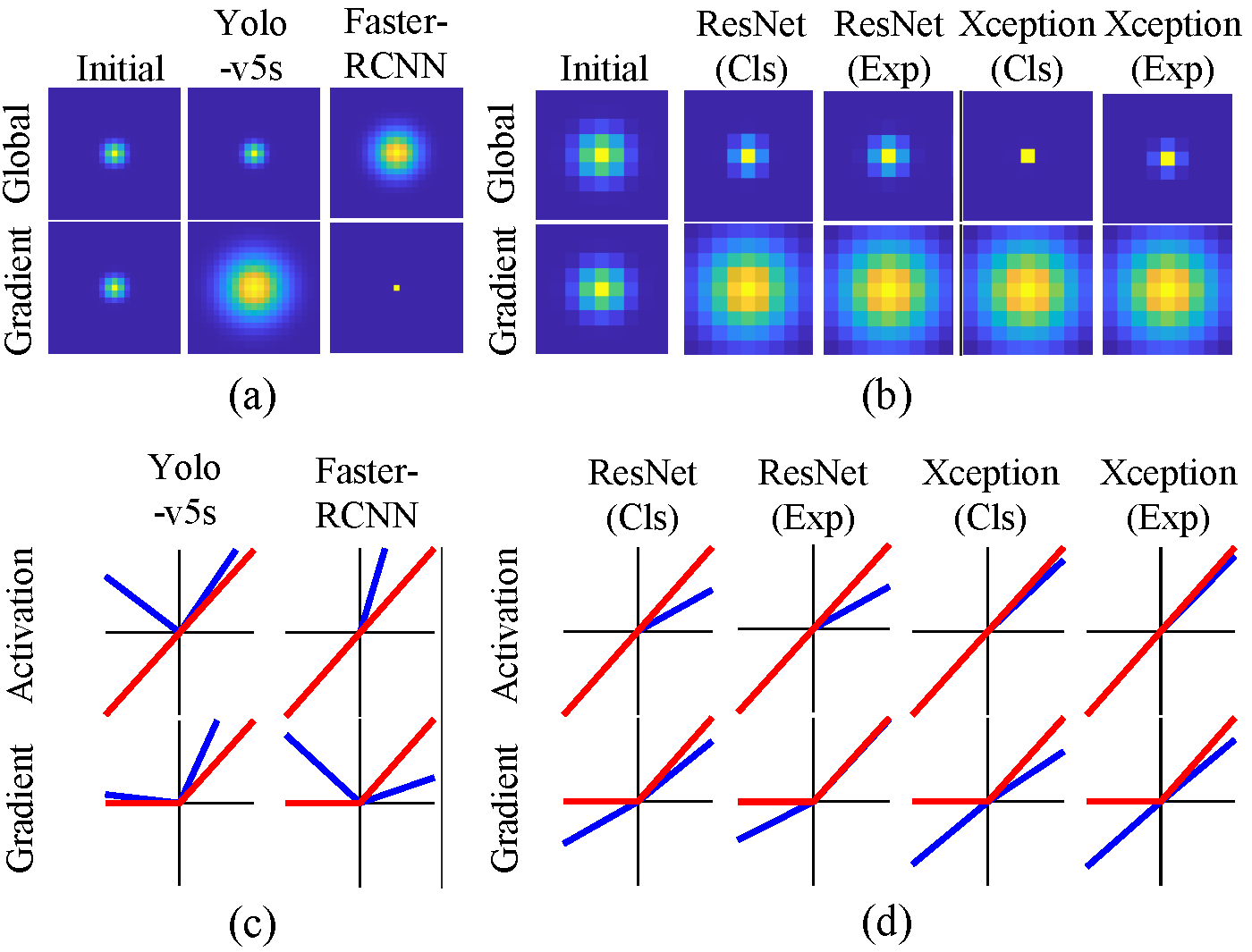}
\caption{Weight visualization for the computer vision models. (a) and (b) are the average Gaussian kernels for object detection models and image classification models, respectively. (c) and (d) are the activation functions averaged over five folds for object detection models and image classification models, respectively. The left column of (a) and (b) are the untrained Gaussian kernels. The red lines in (c) and (d) are the initialization of the  activation function weights before training, while the blue lines are the trained weights.}
\label{fig_15}
\end{figure}

To better understand how the learned functions enhance the faithfulness and plausibility of XAI saliency maps, we visualized the learnable parameters since they are fully interpretable. As shown in Fig. 15a, for Yolo-v5s, a larger smooth kernel was needed for the gradient term. This is because we used the last convolutional layer of the whole model to generate the saliency map, and the salient area was relatively small in the raw gradient term according to the backpropagation algorithm. In contrast, the global smoothing kernel seemed unnecessary, suggesting that the activation map already matched human attention strategies well. In contrast, for Faster-RCNN, the smoothing kernel for gradient was unnecessary since the salient area in gradient was already large when the last convolutional layer of the backbone was used. However, a larger smoothing kernel was required to smooth the grid-like patterns generated by the ROI pooling layer in the global saliency map. As shown in Fig. 15c, the negative parts of the activations and gradients were turned positive in both models, suggesting that these negative parts (i.e., counterfactual information) also play an important role in explanation. Note that for Yolo-v5s, the outputs of the last convolutional layer were activated by a leaky-ReLu function with a leaky factor of 0.1, while a vanilla ReLu activation function was applied to the outputs of the last convolutional layer of Faster-RCNN. Therefore, there existed negative values in activations of Yolo-v5s but not in Faster-RCNN. Together these visualization results suggested that HAG-XAI provided model-specific guidance for generating XAI saliency maps with high plausibility and faithfulness, with the learning functions well generalizable to other detection tasks using a different database with the same models. Fig. 15b illustrates the average Gaussian kernels learned from image classification models with different human-based tasks, and the kernel changes are similar to Yolo-v5s: a larger Gaussian kernel is needed for gradient term while the Gaussian kernel seems useless for global smoothing. From Fig. 15d, a consistent phenomenon can be observed that the negative gradient is useful to enhance the plausibility.

\subsection{Ablation Studies}
\subsubsection{HAG-XAI components}
In HAG-XAI, three functions—a Gaussian smoothing function, a learnable activation function, and an area-based normalization function—were designed to improve faithfulness and plausibility. To examine the effectiveness of these functions, ablation studies were conducted with object detection models. As shown in Tables IV and V, For Yolo-v5s, the XAI method with all three functions achieved the best faithfulness and plausibility. The Gaussian smoothing function contributed the most; without the function, the faithfulness and plausibility significantly dropped. For Faster-RCNN, the XAI method with all three functions achieved the best plausibility, and the Gaussian smoothing function played the most important role in the faithfulness (using insertion increase; the deletion method was affected by the noisy patterns in the gradients and activations, which may lead to incorrect estimations).

\begin{table}[]
\centering
\caption{Ablation study for Yolo-v5s}
\label{tab:Table_4}
\resizebox{\columnwidth}{!}{
\begin{tabular}{ccccccc}
\hline
\multicolumn{3}{c}{\textbf{Function}} & \multicolumn{2}{c}{\textbf{Faithfulness}} & \multicolumn{2}{c}{\textbf{Similarity}} \\ 
\textit{\textbf{G}} & \textit{\textbf{$\varphi$}} & \textit{\textbf{$\mu$}} & \textbf{d-AUC} & \textbf{i-AUC} & \textbf{PCC} & \textbf{RMSE} \\ \hline
× & × & × & 0.4482 & 0.0099 & 0.3550 & 0.0914 \\
$ \checkmark $ & × & × & 0.5612 & 0.0805 & 0.6683 & 0.0900 \\
× & $ \checkmark $ & × & 0.3907 & 0.0019 & 0.2947 & 0.0986 \\
× & × & $ \checkmark $ & 0.3916 & 0.0021 & 0.3140 & 0.0944 \\
$ \checkmark $ & $ \checkmark $ & × & 0.5474 & 0.0793 & 0.6665 & 0.0871 \\
× & $ \checkmark $ & $ \checkmark $ & 0.5330 & 0.0000 & -0.3150 & 0.9643 \\
$ \checkmark $ & × & $ \checkmark $ & 0.5618 & 0.0822 & 0.6873 & 0.0812 \\
$ \checkmark $ & $ \checkmark $ & $ \checkmark $ & \textbf{0.5662} & \textbf{0.0843} & \textbf{0.6910} & \textbf{0.0800} \\ \hline
\end{tabular}}
\end{table}

\begin{table}[]
\centering
\caption{Ablation study for Faster-RCNN}
\label{tab:Table_5}
\resizebox{\columnwidth}{!}{%
\begin{tabular}{ccccccc}
\hline
\multicolumn{3}{c}{\textbf{Function}} & \multicolumn{2}{c}{\textbf{Faithfulness}} & \multicolumn{2}{c}{\textbf{Similarity}} \\ 
\textit{\textbf{G}} & \textit{\textbf{$\varphi$}} & \textit{\textbf{$\mu$}} & \textbf{d-AUC} & \textbf{i-AUC} & \textbf{PCC} & \textbf{RMSE} \\ \hline
$ \times $ & $ \times $ & $ \times $ & \textbf{0.7760} & 0.0562 & 0.5352 & 0.0825 \\
$ \checkmark $ & $ \times $ & $ \times $ & 0.6685 & \textbf{0.1318} & 0.6400 & 0.1097 \\
$ \times $ & $ \checkmark $ & $ \times $ & 0.7614 & 0.0544 & 0.5396 & 0.0822 \\
$ \times $ & $ \times $ & $ \checkmark $ & 0.7676 & 0.0567 & 0.5462 & 0.0823 \\
$ \checkmark $ & $ \checkmark $ & $ \times $ & 0.6831 & 0.1259 & 0.6488 & 0.0995 \\
$ \times $ & $ \checkmark $ & $ \checkmark $ & 0.7561 & 0.0552 & 0.5555 & 0.0817 \\
$ \checkmark $ & $ \times $ & $ \checkmark $ & 0.6767 & 0.1312 & 0.7180 & \textbf{0.0756} \\
$ \checkmark $ & $ \checkmark $ & $ \checkmark $ & \textbf{0.6702} & \textbf{0.1286} & \textbf{0.7198} & \textbf{0.0760} \\ \hline
\end{tabular}%
}
\end{table}

\subsubsection{Effect of selected layer}

\begin{figure}[!t]
\centering
\includegraphics[width=3.2in]{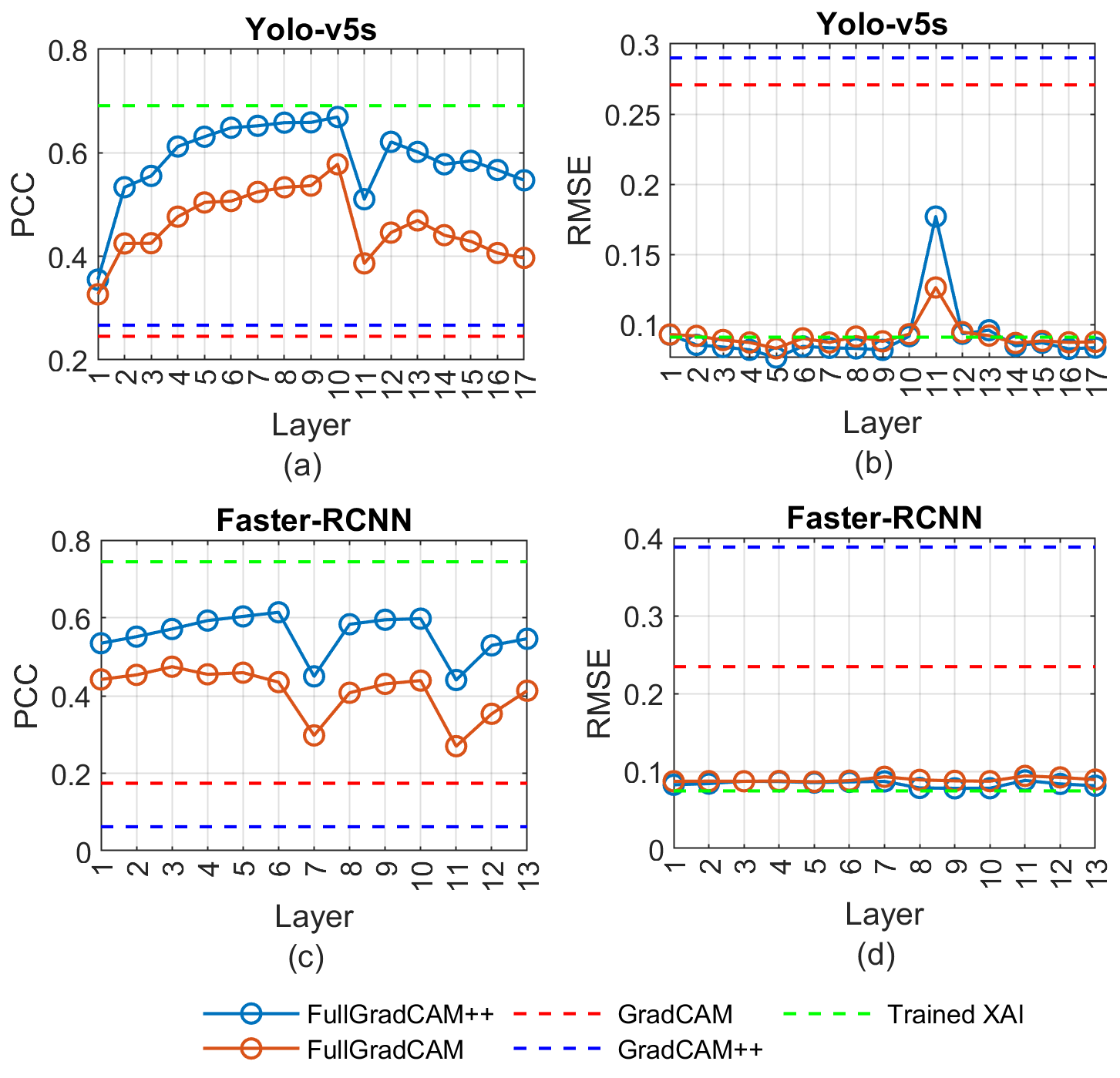}
\caption{The similarity to human attention map (plausibility, as measured in PCC and RMSE) obtained from different XAI methods and different layers. (a) and (b) are for Yolo-v5s, while (c) and (d) are for Faster-RCNN.}
\label{fig_16}
\end{figure}

\begin{figure}[!t]
\centering
\includegraphics[width=3.2in]{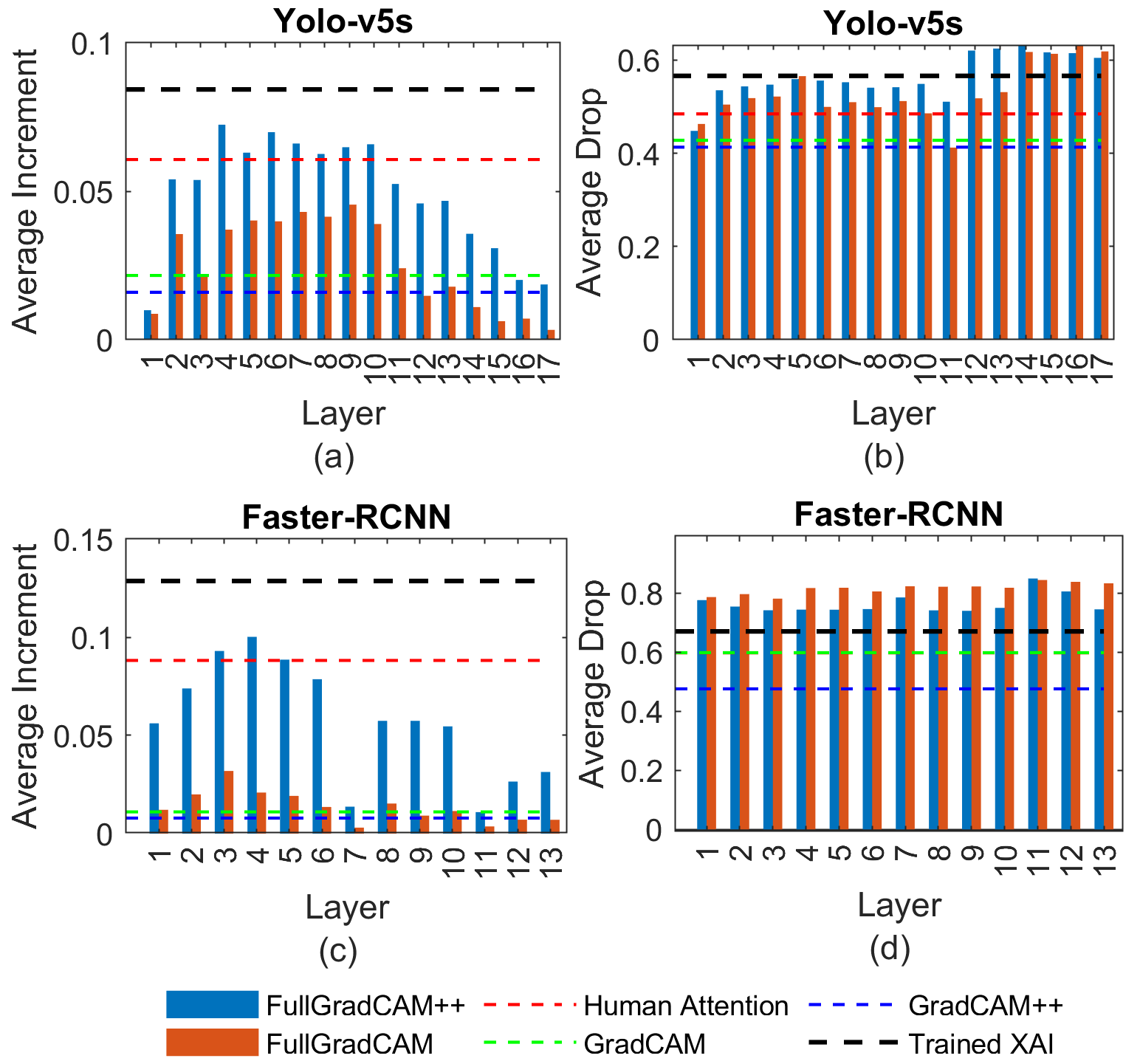}
\caption{The faithfulness measures obtained from different XAI methods and different layers. (a) and (b) are for Yolo-v5s, while (c) and (d) are for Faster-RCNN.}
\label{fig_17}
\end{figure}

Although previous studies suggested that the last convolutional layer should be used to compute the saliency map, the conclusions were mostly drawn on image classification models. Indeed, according to the backpropagation algorithm, shallower layers have a larger salient area in gradients, but the activations may not be as informative as those in deeper layers. Here we generate saliency maps from different layers and evaluate their faithfulness and plausibility for object detection models. The last convolutional layers in each functional block, such as the bottleneck block, are assessed for both Yolo-v5s and Faster-RCNN models. Fig. 16 illustrates the plausibility of different XAI methods in different layers. The results show that although the plausibility as measured in PCC from deeper layers is generally higher than some shallower layers, it is still lower than our trained HAG-XAI. The RMSE values from the deeper layers may be slightly lower due to the small salient regions in the generated saliency map. Similarly, the faithfulness measures demonstrated in Fig. 17 indicate that the proposed HAG-XAI method has higher faithfulness than other XAI methods in any layer. Considering the grid-patterns mentioned before, the deletion score for all layers in Faster-RCNN and the shallow layers in Yolo-v5s can be neglected.

\subsubsection{Effect of degradation and occlusion}

\begin{table*}[]
\centering
\caption{The plausibility and faithfulness under different conditions}
\label{tab:PlauAndFaithUnderDiffCond}
\resizebox{\textwidth}{!}{%
\begin{tabular}{cccccccc}
\hline
\textbf{Measures} & \textbf{Model} & \textbf{Conditions} & \textbf{FGC*} & \textbf{FGC} & \textbf{GC*} & \textbf{GC} & \textbf{HAG-XAI} \\ \hline
\multirow{4}{*}{Plausibility (PCC)} & \multirow{2}{*}{Yolo-v5s} & Occlusion (Y/N) & 0.349/0.374 & 0.317/0.357 & 0.242/0.256 & 0.258/0.295 & 0.691/0.69 \\
 &  & Degradation (Y/N) & 0.345/0.377 & 0.318/0.344 & 0.262/0.209 & 0.284/0.228 & 0.708/0.654 \\
 & \multirow{2}{*}{Faster-RCNN} & Occlusion (Y/N) & 0.517/0.592 & 0.425/0.496 & 0.168/0.199 & 0.049/0.109 & 0.729/0.793 \\
 &  & Degradation (Y/N) & 0.543/0.518 & 0.461/0.401 & 0.212/0.094 & 0.065/0.059 & 0.764/0.702 \\
\multirow{4}{*}{Faithfulness (i-AUC)} & \multirow{2}{*}{Yolo-v5s} & Occlusion (Y/N) & 0.011/0.007 & 0.009/0.007 & 0.025/0.012 & 0.020/0.004 & 0.099/0.037 \\
 &  & Degradation (Y/N) & 0.010/0.011 & 0.008/0.011 & 0.024/0.016 & 0.015/0.018 & 0.080/0.093 \\
 & \multirow{2}{*}{Faster-RCNN} & Occlusion (Y/N) & 0.060/0.044 & 0.013/0.008 & 0.014/0.001 & 0.009/0.005 & 0.128/0.132 \\
 &  & Degradation (Y/N) & 0.055/0.059 & 0.010/0.016 & 0.012/0.009 & 0.007/0.010 & 0.123/0.140 \\ \hline
\multicolumn{8}{l}{\textit{Note: FGC*: FullGrad-CAM++. FGC: FullGrad-CAM. GC*: Grad-CAM++. GC: Grad-CAM.}}
\end{tabular}%
}
\end{table*}

\begin{figure*}[!t]
\centering
\includegraphics[width=7.0in]{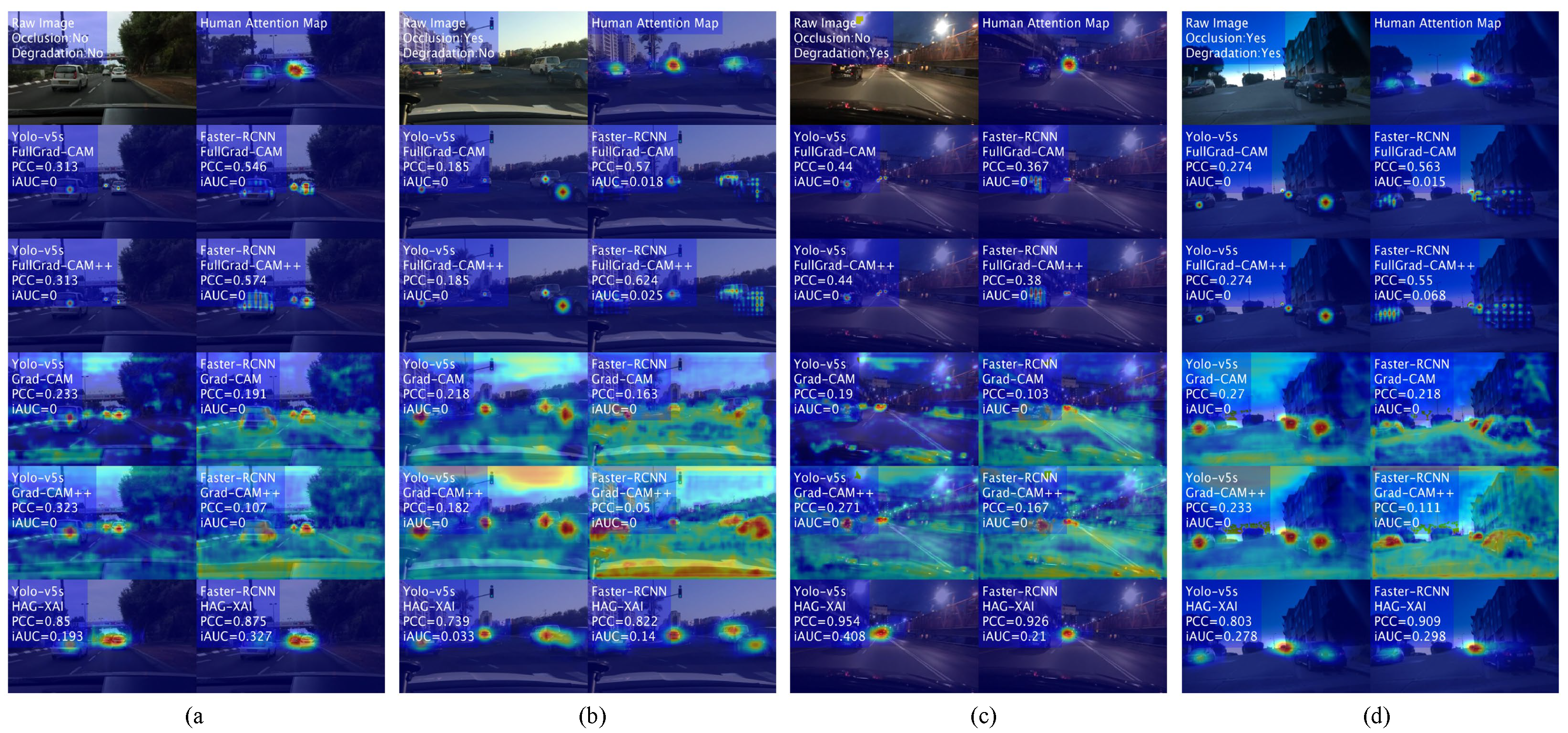}
\caption{Examples of image-level faithfulness and plausibility in different rated conditions: (a) normal; (b) with occlusion; (c) with degradation; (d) with both occlusion and degradation.}
\label{fig_18}
\end{figure*}

To further investigate how different object presentation conditions, such as occlusion or degradation, could influence the performance of the HAG-XAI, we classified each image into one out of four conditions from different combinations of two image properties—with occlusion or not, and with degradation or not. The condition of each image was rated according to the majority choice of three human raters. Images having any vehicle object occluded by any other object were rated as occlusion conditions, whereas images with any vehicle object whose feature identification was influenced by night vision, motion blur, uneven illumination, shadow, or light reflection were rated as degradation conditions. In total, 110 images in dataset A were rated as with degradation conditions, while 122 images were rated as with occlusion conditions. The plausibility and faithfulness of the resulting saliency maps for the two object detection models, two image properties, and five different XAI methods are illustrated in Table VI. Fig. 18 shows example images in degradation or occlusion conditions and their image-level plausibility and faithfulness. As can be seen, our HAG-XAI method achieves the best plausibility and faithfulness performance in every image condition, suggesting the robustness of the proposed method. From the results of HAG-XAI, an interesting phenomenon can be observed: the plausibility scores of degraded images are consistently higher than those of non-degraded images for both models, which means AI models and humans may have more similar attention strategies in identifying degraded than non-degraded images (Fig. 18). The independent t-test shows a significant difference between degraded and non-degraded images in the results from Yolo-v5s ($p$ = 0.04) and a marginal difference in the results from Faster-RCNN ($p$ = 0.09). In contrast, the faithfulness of images in the non-degradation condition did not differ significantly from the degradation condition. These results also indicate that the HAG-XAI may generate more human-like saliency maps in degraded conditions. Another finding is that the average plausibility from the Faster-RCNN model can reach around 0.8 under the non-occlusion condition, which  was marginally higher than that in the occlusion condition ($p$ = 0.06). This result suggests that Faster-RCNN combined with HAG-XAI has the potential to be used as an excellent human attention imitator, especially for non-occluded images. Note that the similarity of HAG-XAI’s output to human attention maps is constrained by the design of the AI model used. In other words, using a model with a cognitively-plausible design may allow the method to generate more human-like attention maps. Future work may examine this possibility.

\section{Conclusion}
In this paper, we first proposed FullGrad-CAM and FullGrad-CAM++ methods for providing saliency-based explanations for object detection models by modifying gradient-based XAI methods originally proposed for image classification models. The methods were then tested and compared with human attention from vehicle detection tasks. Interestingly, while  the resulting salience maps from the methods can achieve higher faithfulness and plausibility for explaining object detection models than traditional gradient-based XAI methods, they fall short in faithfulness as compared with attention maps from humans performing the same object detection tasks. This phenomenon was not observed in image classification models. This result suggests that, in object detection, humans adopt similar attention strategies as AI models, but the current saliency (gradient) based XAI methods are insufficient in capturing these strategies. This motivates us to propose the HAG-XAI method to use human attention as guidance to search for a combination of scale weights and aggregation of features across different gradient and activation maps that can better capture object detection strategies. While for image classification models, HAG-XAI enhances plausibility of the saliency map explanations at the expense of faithfulness, for object detection models it significantly enhances both plausibility and faithfulness simultaneously. Cross-database experimental results indicate that the trained parameters are transferable to other object detection tasks, indicating HAG-XAI’s great generalization ability. HAG-XAI may also be used as a human attention imitator for object detection tasks. Our work thus has significantly enhanced our understanding of the difference between gradient-based methods for explaining image classification and object detection models, and how human attention can be used to enhance both plausibility and faithfulness of XAI methods. Since we found in Study 1 that more human-like explanations can lead to more faithful explanations, Study 2 mainly considered the features matching human attention. However, some non-human-like features may also contribute to faithful explanations. Therefore, we will study exploring non-human-like features to improve the faithfulness of explanations in future works. Another future work may be combining feature maps from multiple layers of the model to allow HAG-XAI to learn to extract finer-grained attention information for a given task to further enhance its performance in providing explanations with high faithfulness and plausibility.

% \section*{Acknowledgments}
% This should be a simple paragraph before the References to thank those individuals and institutions who have supported your work on this article.


\bibliographystyle{IEEEtran}
\bibliography{MEL_final_1}

% \newpage

 

% \end{IEEEbiography}



\vfill

\end{document}